\documentclass[runningheads]{llncs}

\usepackage{amsmath,amssymb} 
\usepackage{xspace}
\usepackage{pifont}

\usepackage{color}
\usepackage[utf8]{inputenc} 
\usepackage[T1]{fontenc}    
\usepackage{hyperref}       
\usepackage{url}            
\usepackage{booktabs}       
\usepackage{amsfonts}       
\usepackage{nicefrac}       
\usepackage{microtype}      

\usepackage{graphicx}
\usepackage{color}

\usepackage{times}
\usepackage{epsfig}
\usepackage{comment}
\usepackage{wrapfig}
\def\onedot{.\xspace}

 \def\eg{e.g\onedot} 
 \def\ie{i.e\onedot}

 \def\etal{\textit{et~al}\onedot}

  \usepackage{multirow}

\newcommand{\marcus}[1]{\textcolor{blue}{#1}}
\newcommand{\tinne}[1]{\textcolor{green}{#1}}

 \newcommand{\fact}[1]{$\langle$#1$\rangle$}
 \newcommand\Mark[1]{\textsuperscript#1}
\begin{document}

\title{Exploring the Challenges towards \\Lifelong Fact Learning} 
\titlerunning{Lifelong Fact Learning} 

\author{Mohamed Elhoseiny\Mark{1}, Francesca Babiloni\Mark{2}, Rahaf Aljundi\Mark{2}, Marcus Rohrbach\Mark{1}, Manohar Paluri\Mark{1}, Tinne Tuytelaars\Mark{2}}
%

\authorrunning{Mohamed Elhoseiny et al.} 


\institute{\Mark{1}Facebook AI Research,  \Mark{2}KU Leuven,  \\\email{\{elhoseiny,mano,mrf\}@fb.com}
 \email{\{francesca.babiloni,rahaf.aljundi,tinne.tuytelaars\}@esat.kuleuven.be}
}
\maketitle

\vspace{-2mm}
\begin{abstract}
So far life-long learning (LLL) has been studied in relatively small-scale and relatively artificial setups. 
Here, we introduce a new large-scale alternative. 
What makes the proposed setup more natural and closer to human-like visual systems is threefold: First, we focus on concepts (or {\em facts}, as we call them) of varying complexity, ranging from single objects to more complex structures such as objects performing actions, and objects interacting with other objects. Second, as in real-world settings, our setup has a long-tail distribution, an aspect which has mostly been ignored in the LLL context. Third, facts across tasks may share structure (e.g., \fact{person, riding,wave} and \fact{dog, riding, wave}). 
Facts can also be semantically related (e.g., ``liger'' relates to seen categories like ``tiger'' and ``lion''). 
Given the large number of possible facts, a LLL setup seems a natural choice. To avoid model size growing over time and to optimally exploit the semantic relations and structure, we combine it with a visual semantic embedding instead of discrete class labels.
We adapt existing datasets with the properties mentioned above into new benchmarks, by dividing them  semantically or randomly into disjoint tasks. This leads to two large-scale benchmarks with 906,232 images and 165,150 unique facts, 
on which we evaluate and analyze  state-of-the-art LLL methods. 
\end{abstract}

\vspace{-5mm}
\section{Introduction}
\label{sec:introduction}

Humans can learn new visual concepts without significantly forgetting previously learned ones and without necessarily having to revisit previous ones.
In contrast, the majority of existing artificial visual deep learning systems assume a replay-access to all the training images and all the concepts during the entire training phase 
-- e.g., going a large number of epochs over the 1000 classes of ImageNet. 
This assumption also applies to systems that learn concepts by reading the web (e.g.,  
\cite{mitchell2015never,chen2013neil,divvala2014learning})
or that augment CNNs with additional units to better transfer knowledge to new tasks such as \cite{wang2017growing}. 

To get closer to human visual learning and to practical application scenarios, where data often cannot be stored due to physical restrictions (e.g. robotics) or policy (e.g. privacy), the scenario of 
lifelong learning (LLL)  has been proposed. 
The assumption of LLL  
is that only a subset of the concepts and corresponding training instances are available at each point in time during training. 
%
Each of these subsets is referred to as a ``task'', originating
from robotics applications 
\cite{thrun1998clustering}.
This leads to a chain of learning tasks trained on a time-line. 
 While training of the first task is typically unchanged, the challenge is how to train the remaining tasks without reducing performance on the earlier tasks. Indeed, when doing so naively, e.g. by fine-tuning previous models,
this results in what is known as {\em catastrophic forgetting}, i.e.,  the accuracy on the earlier tasks drops significantly. Avoiding such catastrophic forgetting is 
the main challenge addressed in the lifelong learning literature. 
\begin{figure}[t]
\centering
\vspace{-3mm}
\begin{minipage}{0.52\textwidth}
\centering
\includegraphics[width=1.0\textwidth]{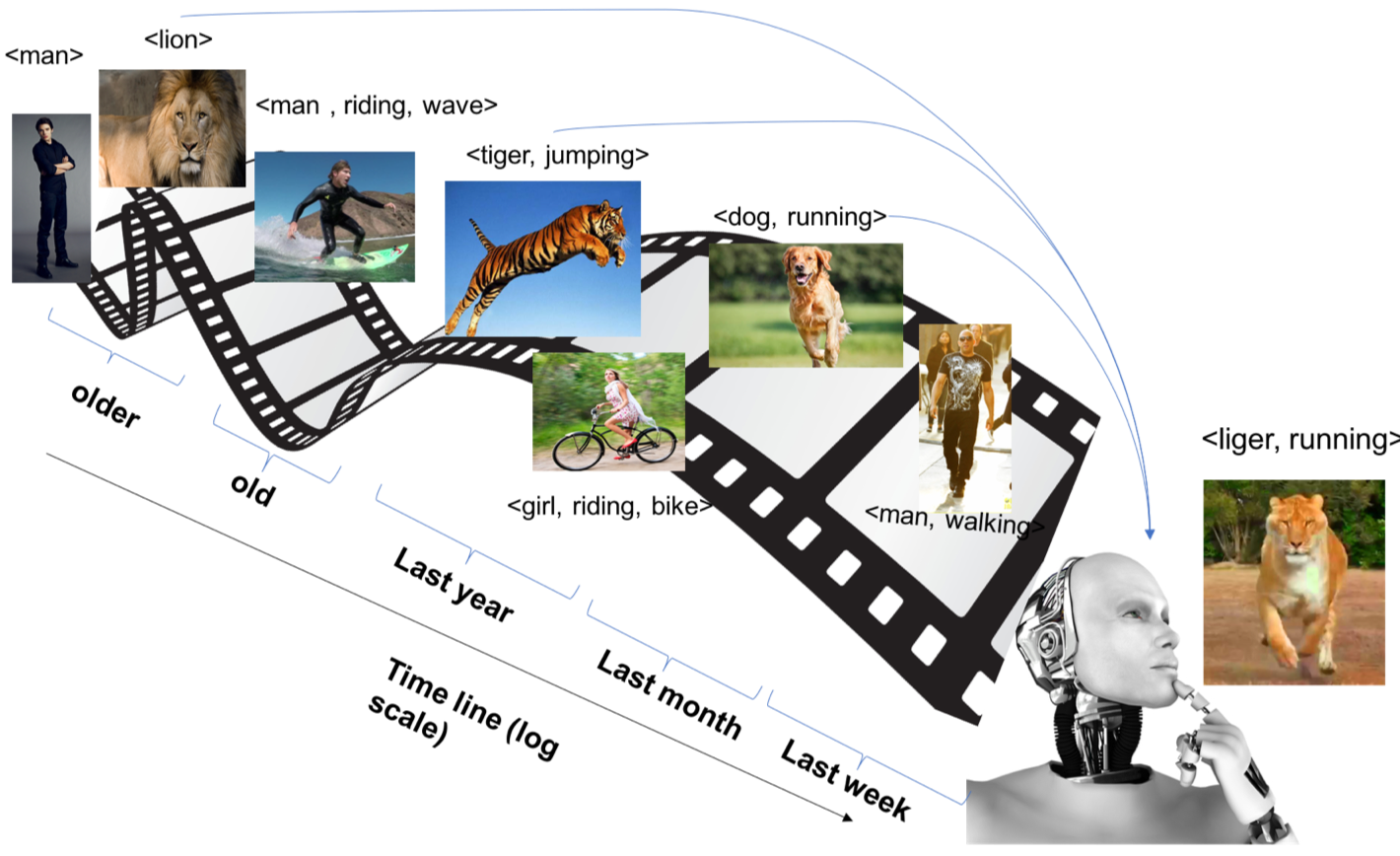}
   \caption{Lifelong  Fact Learning}
       \label{fig1_teaser}

\end{minipage}
\begin{minipage}{0.46\textwidth}
\centering
    \includegraphics[width=1.0\textwidth]{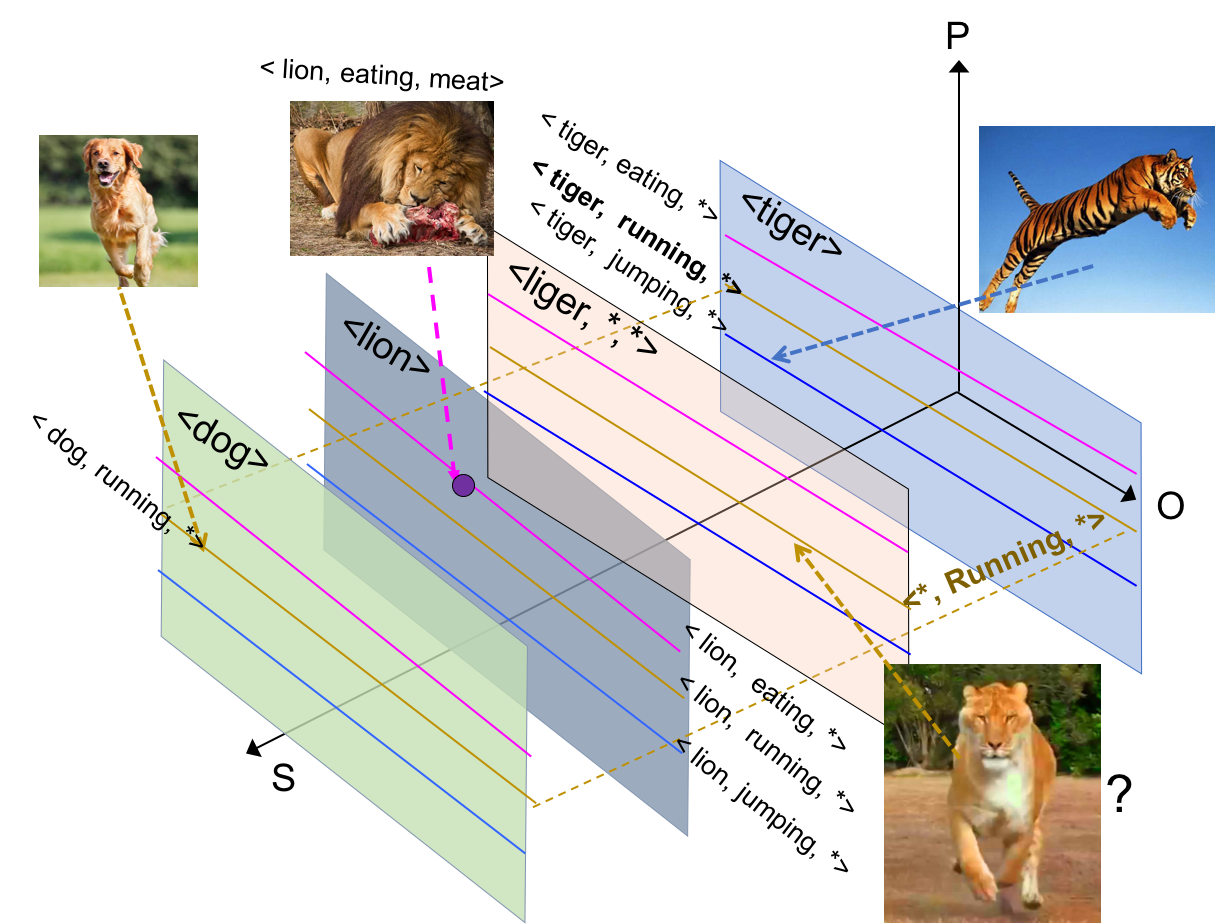}
      \caption{ Structured Fact Representation}
      \label{fig2_structuredspace}
\end{minipage}
\vspace{-4mm}
\end{figure}


%
%
%
\noindent  \textbf{Lifelong Fact Learning (LLFL). } 
Existing works on LLL have focused 
{mostly}
on image classification tasks (\eg 
\cite{aljundi2016expert,kading2016fine,li2016learning,rebuffi2016icarl,triki2017encoder,zenke2017improved}), in a relatively small-scale and somewhat artificial setup. 
A sequence of tasks is defined, either by combining multiple datasets (e.g., learning to recognize MITscenes, then CUB-birds, then Flowers), by dividing a dataset (usually CIFAR100 or MNIST) into sets of disjoint concepts, or by permuting the input (permuted MNIST).  {Instead, in this work we propose a LLL setup with the following more realistic and  desirable learning characteristics:} 

\noindent
\textit{{\bf 1.} Long-tail:} Training data can be highly unbalanced with the majority of concepts occurring only rarely, which is in contrast to many existing benchmarks (e.g., ~\cite{he2016deep,krizhevsky2012imagenet,simonyan2014very}).

\noindent
\textit{{\bf 2.} Concepts of varying complexity:}
We want to learn diverse concepts, including not only objects but also actions, interactions, attributes, 
as well as combinations thereof. 

\noindent
\textit{{\bf 3.} Semantic and structure aware:} We 
want to connect semantically related visual facts. For example, if we have learned ``lion'' and ``tiger'' earlier, that 
can help us later in time to learn a ``liger'' (a rare hybrid cross between a male lion and a female tiger), even with just a few examples. Relating this to point (2) above,  this further allows {\em compositional lifelong learning} to help recognize  new facts (\eg \fact{dog, riding, wave}) based on facts seen earlier in time (\eg \fact{person, riding, wave}  and \fact{girl, walking, dog}).

To the best of our knowledge, none of the existing LLL literature explored these challenges. 
We denote studying lifelong learning with the 
aforementioned characteristics as {\em lifelong fact learning} (LLFL); see Fig.~\ref{fig1_teaser}.   


\noindent  \textbf{A Note on Evaluation Measures.} We argue that the evaluation of LLL methods should be reconsidered. In the standard LLL (with a few notable exceptions, such as~\cite{aljundi2016expert,chaudhry2018riemannian}), the trained models are judged by their capability to recognize each task's categories individually assuming the absence of the categories covered by the remaining tasks -- \textit{not necessarily realistic}. 
Although 
the performance of each task in isolation
is an important characteristic, it might be deceiving. Indeed, a learnt representation could be good to classify an image in a restricted concept space covered by a single task, but may not be able to classify the same image when considering all concepts across tasks. It is therefore equally important to measure the ability to distinguish the learnt concepts across all the concepts over all tasks. This is important since the objective of LLL is to model the understanding of an ever growing set of concepts over time. 
%
\textit{To better understand how  LLL performs in real world conditions, we advocate evaluating the existing methods across different tasks}. We named that evaluation {\em Generalized lifelong learning} (G-LLL), in line with the idea of Generalized zero-shot learning proposed in~\cite{chao2016empirical}. We detail the evaluation metric in Sec. \ref{sec:evalmetrics}. 

\noindent  
\textbf{Advantages of a Visual-Semantic Embedding.}
As illustrated in 
Fig.~\ref{fig1_teaser}, 
we expect to better understand \fact{liger, running} by leveraging previously learnt  facts such as \fact{lion},  \fact{tiger, jumping} and \fact{dog, running}. 
This shows how both semantics and structure are helpful for understanding.
To our knowledge, such semantic awareness has not  been studied in a LLL context.  To achieve this, we use a visual-semantic embedding model where semantic labels  and images are embedded in a joint space. For the semantic representation,  we leverage semantic external knowledge using word embeddings -- in particular word2vec~\cite{mikolov2013distributed}. 
These word embeddings were shown to
efficiently learn semantically meaningful floating point {vector} representations of {words}. 
For example, the average vector of lion and tiger is closest to liger. 
This can help semantically similar concepts to  learn better from one another, as shown in~\cite{zhang2018large,elhoseiny2017sherlock} in non LLL scenarios. 
Especially in our long-tail setting, this can be advantageous. 
Additionally, by working with an embedding instead of discrete concept labels as in~\cite{fernando2017pathnet,lee2017overcoming,kirkpatrick2016overcoming,zenke2017improved}, we avoid that the model keeps growing 
as new concepts get added, which would make the model 
less scalable and limit the amount of sharing.
%
%


\noindent\textbf{Contributions.}
First, we introduce a midscale and a large scale benchmark for Lifelong Fact Learning (LLFL), with two splits each, a random and a semantic split. Our approach for creating a semantically divided benchmark is general and could be applied similarly to other datasets or as more data becomes available.
Second, we advocate to focus on a more generalized evaluation (G-LLL)  
where test-data cover the entire label space across tasks.
Third, we evaluate existing LLL approaches in both the standard and the generalized setup on our new LLFL benchmarks.
Fourth,  we discuss  the limitations of the current generation of LLL methods in this context, which forms a basis for advancing the field in future research.
  Finally, 
this paper aims to answer the following questions: {\em How do existing LLL methods perform on a large number of concepts? 
What division of tasks is more helpful to continually learn facts at scale (semantically divided vs randomly divided)?
How does the long-tail distribution of the facts limit the performance of the current methods?} 

\vspace{-2mm}
\section{Related Work}
\vspace{-3mm}
\label{sec:related}
\newcommand{\xmark}{\ding{55}}%
\newcommand{\cmark}{\ding{51}}
\begin{table}[t!]
\centering
\vspace{-3mm}
\scalebox{0.6}
{
\begin{tabular}{|l|c|c|c|c|c|c|}
\hline
Dataset & Structured/Diverse & Long-Tail & Classes &Examples & Task Count& Split Type \\\hline
MNIST & \xmark & \xmark &  10 & 60000 & 2 to 5& R \\ \hline 
CIFAR (used in~\cite{rebuffi2016icarl,zenke2017improved,lopez2017gradient}) & \xmark & \xmark & 100 & 60000  & 2  and 5 & R\\  \hline  
ImageNet and CUB datasets (used  in~\cite{lee2017overcoming})& \xmark & \xmark &  1200 &1211000  & 2  & R \\ \hline 
 Scenes, CUB, VOC, and  Flowers (used in ~\cite{li2016learning,aljundi2016expert,triki2017encoder})& \xmark & \xmark &  122-526 &5908-1211000  & 2 & S  \\ \hline 
{8 Dataset Sequence~\cite{mas}}&  \xmark & \xmark  & 889 & 714387&  8 & S \\ \hline 
CORe50 ~\cite{lomonaco17corl} / iCUBWorld-Transf(~\cite{Pasquale2016IROS} & \xmark & \xmark   & 10 (50)/15(150) & 550/900 sessions & 10 & S\\    \hline   
\textbf{Our Mid-Scale LLFL Benchmark } & \cmark & \xmark &186 & 28624 & 4 & S $\&$ R\\ \hline 
\textbf{Our Large Scale LLFL  Benchmark } & \cmark & \cmark &165150 & 906232 & 8 & S $\&$ R\\ \hline 
\end{tabular}
}
\caption{Comparison of some existing Task Sequences. Split Type is either S (Semantic), R (Random), or S$\&$R (Both Semantic and Random splits are provided) 
}
\label{table:datasets}
\vspace{-8mm}
\end{table}




\textbf{Previous Evaluations of LLL. }
In Table~\ref{table:datasets}, we compare some of the popular  datasets/ benchmarks used in LLL. As also noted by Rebuffi \etal~\cite{rebuffi2016icarl}, there is limited agreement about the setup. 
Most build a task sequence by combining or dividing standard object/scene recognition datasets. 
%
In the context of robotics, Lomonaco and Maltoni~\cite{lomonaco17corl} introduced the CORe50 dataset which consists of relatively short  RGB-D video fragments (15sec) of handheld domestic objects. They focus both on category-level as well as instance-level object recognition. With 50 objects belonging to 10 different categories it is, however, relatively small scale and limited in scope. Pasquale~\etal with a similar focus proposed the iCUBWorld-Transf dataset~\cite{Pasquale2016IROS} with 200 real objects divided in 20 categories. For CORe50 and iCUBWorld-Transf, the number of instances is shown in parenthesis  in Table~\ref{table:datasets}. In a reinforcement learning setup,   Kirkpatrick \etal~\cite{kirkpatrick2016overcoming} and Fernando \etal~\cite{fernando2017pathnet}  performed interesting  LLL experiments using a sequence of Atari Games 
as tasks.
In contrast to all of the above, we aim at a more natural and a larger-scale setup; see last two rows in Table~\ref{table:datasets}. Our benchmarks are more structured and challenging, due to the large number of classes and the long-tail distribution.  



\noindent 
\textbf{Existing LLL Approaches. }
LLL works may be categorized into data-based and model-based approaches. In this work, we
do not consider methods that require storing samples from previous tasks in an episodic memory~\cite{rebuffi2016icarl,lopez2017gradient}.   


In \textit{data-based approaches}~\cite{li2016learning,shmelkov17iccv,triki2017encoder}, 
the new task data is used to estimate and preserve the model behavior on previous tasks, mostly via a knowledge distillation loss as proposed in {\em Learning without Forgetting}~\cite{li2016learning}. { These approaches are typically applied to 
a sequence of tasks with different output spaces.} 
To reduce the effect of distribution difference between tasks, Triki~\etal~\cite{triki2017encoder} propose to incorporate  a shallow auto-encoder to further control the changes to the learned features, 
while Aljundi~\etal~\cite{aljundi2016expert} train a model for every task (an expert) 
and use auto-encoders to help determine the most related expert at test time given an example input. 


\textit{Model-based approaches}~\cite{fernando2017pathnet,lee2017overcoming,kirkpatrick2016overcoming,zenke2017improved} on the other hand focus on the parameters of the network. 
%
The key idea is to define an importance weight $\omega_i$ for each parameter $\theta_i$ in the network indicating the importance of this parameter to the previous tasks. 
When training a new task, network parameters with high importance are discouraged from being changed.
In {\em Elastic Weight Consolidation}, Kirkpatrick \etal~\cite{kirkpatrick2016overcoming} estimate the importance weights $\mathbf{\Omega}$ based on 
the inverse of the Fisher Information matrix. 
%
Zenke~\etal~\cite{zenke2017improved} propose {\em Synaptic Intelligence}, an online continual model where $\mathbf{\Omega}$ is defined by the contribution of each parameter to the change in the loss,
and weights are  accumulated for each parameter during training.
{\em Memory Aware Synapses}~\cite{mas}  measures $\mathbf{\Omega}$ by the effect of a change in the parameter to the function learned by the network, rather than to the loss. This allows to estimate the importance weights not only in an online fashion but also  without the need for labels. 
Finally,  {\em Incremental Moment Matching} 
\cite{lee2017overcoming} 
is a scheme to merge models trained for different tasks.
Model-based methods seem particularly well suited for our setup, given that we work with an embedding instead of disjoint output spaces. 

\vspace{-2mm}
\section{Our Lifelong Fact Learning Setups}
\vspace{-2mm}
\label{sec:benchmark}




We aim to build two LLL benchmarks that consist of a diverse set of facts (two splits for large-scale and two splits for mid-scale).
The benchmarks capture different types of facts including objects (e.g., \fact{lion}, \fact{tiger}), objects performing some activities (e.g., \fact{tiger, jumping},   \fact{dog,running}), and interactions between objects (e.g., \fact{lion, eating, meat}). Before giving details on the benchmark construction, we first explain how we represent facts.

\textbf{A visual-semantic embedding for facts.} 
Inspired by~\cite{plummer2017phrase,elhoseiny2017sherlock}, we represent every fact for our LLL purpose 
by three pieces represented in a semantic  continuous space. $\textbf{S} \in \mathbb{R}^{d}$  represents object or scene  categories. 
$\textbf{P}\in \mathbb{R}^{d}$ represents predicates, \eg  actions or interactions.
$\textbf{O}\in \mathbb{R}^{d}$ represents objects that interact with $\textbf{S}$. Each of $\textbf{S}$, $\textbf{P}$, and $\textbf{O}$ lives in a high dimensional semantic space. 
By concatenating these three representations, we obtain
a structured space that can represent all the facts that we are interested to study in this work. Here, we follow \cite{elhoseiny2017sherlock} and semantically represent each of $\textbf{S}$, $\textbf{P}$, and $\textbf{O}$  by their corresponding word2vec embeddings~\cite{mikolov2013distributed}.
\begin{equation}
\small
\begin{split}
\textbf{\fact{S,P,O} \text{(e.g., <person, riding, horse>)}: } 
\mathbf{t}  = [\mathbf{t}_S, \mathbf{t}_P, \mathbf{t}_O] \\\  
\textbf{\fact{S, P,*} \text{ (e.g., <man, walking, *>)}: } 
\mathbf{t} = [\mathbf{t}_S , \mathbf{t}_P ,  \mathbf{t}_O = *] 
 \\
 \textbf{ \fact{S,*,*}  \text{(e.g., <dog, *, *>)}: }
\mathbf{t} = [\mathbf{t}_S , \mathbf{t}_P = *,  \mathbf{t}_O = *] 
\end{split}
\label{eq3} 
\end{equation}
%
where $[ \cdot, \cdot, \cdot]$ is  the concatenation operation and  $*$ means undefined and set to zeros. The rationale behind this notation convention
is that if a ground truth image is  annotated as  \fact{man},  
this could also be
\fact{man, standing}  or \fact{man, wearing, t-shirt}. Hence, we represent the man as \fact{man, *,*}, where $*$ indicates that we do not know if that ``man'' is doing something. 
Figure~\ref{fig2_structuredspace} shows how different fact types could be represented in this space, with $\textbf{S}$, $\textbf{P}$, and $\textbf{O}$  visualized as a single dimension.  Note that $\textbf{S}$ facts like \fact{lion} 
are represented as a hyper plane in this space. While \fact{tiger, jumping} and \fact{lion, eating, meat} are represented as a hyper-line and a point respectively.
 \begin{figure*}[t!]
    \vspace{-2mm}
  \centering
    \includegraphics[width=0.78\textwidth]{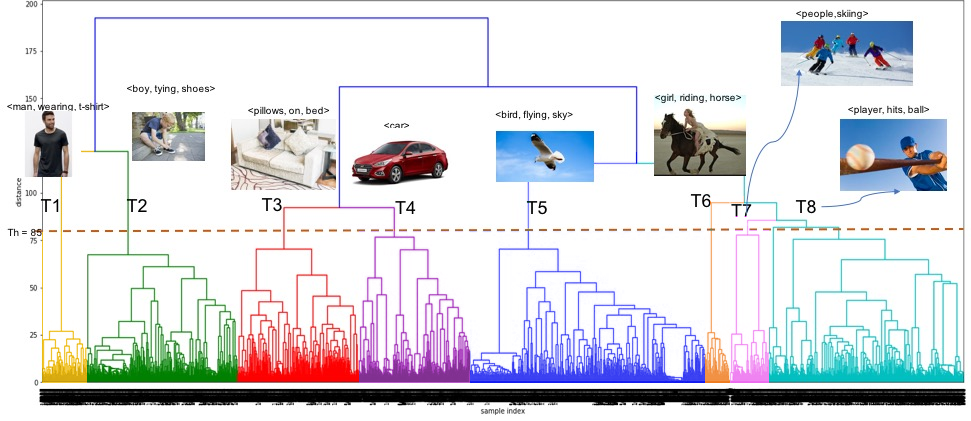}
    \vspace{-2mm}
      \caption{Lifelong Learning Semantically Divided Benchmark: 8 Tasks generated by agglomerative clustering in the semantic space of facts. The method is general and can be re-applied as more images and labels become available.
      }
      \label{fig_benchmark}
      \vspace{-4mm}
\end{figure*}

\vspace{-2mm}
\subsection{Large Scale LLFL Benchmark}
We build our setup on top of the large scale fact learning dataset introduced by~\cite{elhoseiny2017sherlock}, denoted as  {\em Sherlock LSC} (for Large SCale). It has more than 900,000 images and 200,000 unique facts, from which we excluded attributes. 
The dataset was created by extracting facts about images from image descriptions and image scene graphs. It matches our desired properties of being \emph{long-tailed} 
and \emph{semantic-aware} due to its structure.  



Given this very large set of facts and examples for each of them, we want to learn them in a LLL setting. This involves splitting the data into a sequence of {\em disjoint} tasks
(that is, 
with no overlap in the facts learned by different tasks).
However, due to their structured nature, facts may be {\em partially} overlapping across tasks, \eg have the same subject or object.
In fact, we believe that some knowledge reappearing across different tasks is a desired property in many real-life LLL settings, as it facilitates knowledge transfer. 
On the other hand, 
one could argue that the different tasks that real world artificial agents are exposed to, are likely to cover different domains -- a setting more in line with existing LLL works. 
To study both scenarios, we built a semantically divided split (less sharing among tasks) and a randomly divided one (with more sharing). 

\begin{table}[b!]
\centering
\vspace{-2mm}
\caption{Number of Unique Facts (i.e., Labels) and Images in each of the 8 tasks for our Semantically and Randomly Divided Large-Scale Benchmark for  \fact{S}, \fact{S,P}, and \fact{S,P,O}}
\label{table_8tasks}
\vspace{-2mm}
\scalebox{0.6}
{
    \vspace{-2mm}
\begin{tabular}{@{}|c|c|c|c|r|r|r||c|c|c|c|r|r|r|@{}} \hline
\multicolumn{7}{|c||}{\textbf{Random}}  & \multicolumn{7}{c|}{\textbf{Semantic}} \\ \hline
Task & Facts-SPO & Facts-SP & Facts-S & \multicolumn{1}{|l|}{images-SPO} & \multicolumn{1}{|l|}{images-SP} & \multicolumn{1}{|l||}{images-S} & \multicolumn{1}{|l|}{Task} & Facts-SPO & Facts-SP & Facts-S & \multicolumn{1}{|l|}{images-SPO} & \multicolumn{1}{l}{images-SP} & \multicolumn{1}{|l|}{images-S}\\ \hline
1 & 19311 & 7100 & 1114 & 40244 & 41523 & 102605 & 1 & 6577 & 224 & 1 & 19311 & 41523 & 102605\\ \hline
2 & 16051 & 5926 & 961 & 35265 & 34234 & 99442 & 2 & 25552 & 2871 & 3 & 16051 & 34234 & 99442\\ \hline
3 & 14594 & 5305 & 796 & 27812 & 32215 & 58009 & 3 & 12400 & 517 & 250 & 14594 & 32215 & 58009\\ \hline
4 & 13430 & 4851 & 761 & 26069 & 24701 & 66355 & 4 & 7923 & 305 & 46 & 13430 & 24701 & 66355\\ \hline
5 & 8713 & 3255 & 524 & 18217 & 17588 & 100465 & 5 & 42264 & 24381 & 6321 & 8713 & 17588 & 100465\\ \hline
6 & 14125 & 5274 & 830 & 30827 & 32830 & 57656 & 6 & 2819 & 413 & 7 & 14125 & 32830 & 57656\\ \hline
7 & 16688 & 5935 & 876 & 34313 & 30582 & 55362 & 7 & 6917 & 1181 & 4 & 16688 & 30582 & 55362\\ \hline
8 & 13083 & 4845 & 802 & 28255 & 27525 & 366338 & 8 & 11543 & 12599 & 32 & 13083 & 27525 & 366338\\ \hline \hline 
\multicolumn{1}{|l|}{\textbf{Total}} & \multicolumn{1}{|c|}{115995} & \multicolumn{1}{|c|}{42491} & \multicolumn{1}{|c|}{6664} & 241002 & 241198 & 906232 & \multicolumn{1}{|c|}{36} & \multicolumn{1}{|c|}{115995} & \multicolumn{1}{|c|}{42491} & \multicolumn{1}{|c|}{6664} & 115995 & 241198 & 906232\\ \hline
\end{tabular}
}
\label{tbl_lsc_statistics}
    \vspace{-2mm}
\end{table}

\textbf{Large Scale Semantically Divided Split.} 
We
semantically group the facts to create the tasks needed to build our benchmark, \ie we cluster similar facts and assign each cluster to a task $T_i$. 
In particular, we first populate the structured embedding space with all the training facts and then cluster the facts semantically with a custom metric. 
Since our setting allows diverse facts where one or two of the three components might be undefined, we need to consider  a proper similarity measure to allow clustering the facts.  We assume that the structured fact space is Euclidean and has unit norm (\ie, cosine distance). Hence, we define the distance between two facts $\mathbf{t}_i$ and $\mathbf{t}_j$ as follows: 
\begin{equation}
\small
\begin{split}
D(\mathbf{t}^i , \mathbf{t}^j) = & \frac{w_S^{ij}\|\mathbf{t}^i_S-\mathbf{t}^j_S\|^2+w_P^{ij} \|\mathbf{t}^i_P-\mathbf{t}^j_P\|^2 +w_O^{ij} \|\mathbf{t}^i_O-\mathbf{t}^j_O\|^2}{w_S^{ij}+w_P^{ij}+w_O^{ij}}  \\
w_{l}^{ij} =&0\text{ only if }\mathbf{t}^i_l=*\text{ or }\mathbf{t}^j_l=* , l\in\{\textbf{S,P,O}\}\text{ and } w_{l}^{ij}=1 \text{ otherwise}.
\end{split}
\label{eq_sim}
\end{equation}
with $w_l$ 
an indicator value distinguishing between singleton facts, pairs or triplets.
The intuition behind this distance measure is that we do not want to penalize the $*$ (undefined) part when comparing for example $t_i=$\fact{person,*,*} to 
$t_j=$\fact{person,jumping,*}. 
In this case the distance should be 
zero
since the $*$ piece does not contribute to the distance measure. 
We rely on bottom-up hierarchical agglomerative clustering which clusters facts together monotonically based on their distance into disjoint tasks using the aforementioned distance measure. This clustering algorithm recursively merges the pair of clusters that minimally increases a given linkage metric.
In our experiments, we use the nearest point algorithm, i.e. clustering with single linkage.
An advantage of the agglomerative clustering algorithm is that 
the distance measure need not be a metric. 
The result of the clustering is shown in the form of a Dendrogram  in Fig.~\ref{fig_benchmark}.
By looking at the clustered facts, we choose a threshold of 85, shown by the 
\tinne{red-dashed}
line, leading to $n=8$ tasks in our work, as detailed further in Table~\ref{table_8tasks}.
We attach in the supplementary a PCA visualization of the generated tasks using the word embedding representation of each fact and histogram over facts to illustrate the long-tail. 
We note that the number of facts and images is not uniform across tasks, and some tasks are likely easier than others. We believe this mimics  realistic scenarios, where an agent will have to handle tasks which are of diverse challenges.

\textbf{Large Scale Randomly Divided Split.} 
We also introduce a randomly divided benchmark where the facts are divided randomly over tasks rather than based on semantics. The semantic overlap 
between randomly split tasks is expected to be higher than for the semantically-split tasks where the semantic similarity between tasks is minimized. Table~\ref{table_8tasks} shows the task information some further information 
for 
both types of splits.
For the random split, we make sure that the tasks contain a balanced number of facts and of corresponding training and test images by selecting the most balanced candidate out of 100 random trials. Hence,  the random split is more balanced  by construction in terms of training images per task. Since we split the data randomly into tasks, semantically related facts would be distributed across tasks.  

\vspace{-2mm}
\subsection{Mid Scale Balanced LLFL Benchmark}
Compared to the large scale dataset, this dataset is more balanced, with the long-tail effect being less pronounced. This allows us to contrast any change in the behavior of the LLL methods going from a uniform distribution to a long-tail distribution. 
We build the mid-scale LLFL dataset  on top of the 6DS dataset introduced in~\cite{elhoseiny2017sherlock}. It is composed of $186$ unique facts and $28,624$  images, divided in $14,599$ training samples and $14,025$ test samples. 
We 
divided this dataset randomly and semantically into 4 tasks.


{\bf  Mid-Scale Semantic Split. } 
We  use the same  mechanism for clustering as described above to create a benchmark of 4 tasks that are semantically divided.  By visually analyzing the clusters, we find the following distribution:
{\em - Task 1:} facts describing human actions such as \fact{person,riding bike}, \fact{person, jumping}, 
{\em - Task 2:} facts of different objects such as \fact{battingball}, \fact{battingstumps}, \fact{dog}, \fact{car}, 
{\em - Task 3:} facts describing humans holding or  playing musical instruments, such as \fact{person, playing,flute}, \fact{person, 	holding, cello}, etc.
{\em - Task 4:} facts describing human interactions such as \fact{person, arguing with, person}, \fact{person, dancing with, person}.

{\bf Mid-Scale Random Split.}
We followed the same procedure described in the large scale benchmarks to split the facts into 4 different random groups. Note that \cite{mas} evaluated image retrieval (with average precision) on a similar random-split of 6DS \cite{elhoseiny2017sherlock} while in this work we look at the task of fact recognition (measured in accuracy), which is meaningful for both the mid-scale and the large-scale benchmarks (our focus) since the vast majority of the facts has only one image example.  

\vspace{-2mm}
\section{Lifelong Learning Approaches}
\label{sec:methods}
In this section, we first formalize the life-long learning task, then we review the evaluated methods, and finally we explain how we adapt them to fact learning.
\subsection{LLL Task}
Given a training set  $\mathcal{D} = \{ (\mathbf{x}_k, y_k) \}_{k=1}^{M} $, we learn from different tasks $T_1,T_2,\ldots,T_N$ over time where $T_n\subset\mathcal{D}$. $y_k$ in our benchmarks are structured labels. 
For most model-based approaches,  we can formalize the LLL loss as follows. 
The loss of training the new $n^{th}$ task is $L_n(\theta)$, where $\theta$ are the parameters of the network such that $\theta_{i}$ is the $i^{th}$  parameter of  an arbitrary  neural network (a deep neural network with both convolutional and fully connected layers, in our case). $L(\theta)$ is defined as $
L(\theta)=L_n(\theta)+\frac{\lambda}{2} \sum_i \omega^{n-1}_{i}(\theta_{i}-\theta^{n-1}_{i})^2$,
where $\lambda$ is a hyperparameter for the regularizer, $\theta_{i}^{n-1}$ the previous task's network parameters, and $\omega_{i}^{n-1}$ a weight indicating the importance of parameter $\theta_i$ for all tasks up to $n-1$. Hence, we strongly regularize the important parameters at the previous time step (i.e., high $\omega^{n-1}_{i}$) and weak regularization on the non-important parameters (i.e., low $\omega^{n-1}_{i}$). This way, we allow changing the latter more freely.  
Under this importance weight based  framework,  Finetuning, Intelligent Synapses \cite{zenke2017improved} and Memory Aware Synapses \cite{mas} are  special cases.

\subsection{Evaluated Methods}

\noindent \textbf{(1) Finetuning} ({FT}): 
 FT is a common LLL baseline. It does not involve any importance parameters, 
so  $\omega^{n}_{i} =0, \forall i$.

\noindent
\textbf{(2) Synaptic Intelligence~\cite{zenke2017improved}} ({Int.Synapses})
estimates the importance weights in an online manner while training based on the contribution of each parameter to the change in the loss. The more a parameter $\theta_i$ contributes to a change in the loss, the more important it is. 


\noindent
\textbf{(3) Memory  Aware Synapses~\cite{mas}} ({MAS}) 
defines importance of parameters in an online way based on their contribution to the change in the function output. $
\omega_{i}^n =\frac{1}{M_n}\sum_{k=1}^{M_n}  \mid\mid g_{i}(x_k)  \mid\mid$, 
where $g_{i}(x_k)=\frac{\partial (F(x_k; \theta))}{\partial \theta_{i}}$ is the gradient of the learned function  with respect to  $ \theta_{i}$  evaluated at the data point $x_k$.
$F$ maps the input $X_i$ to the output $Y_i$.  This mapping 
is the target that MAS preserves to deal with forgetting.

\noindent
\textbf{(4) ExpertGate~\cite{aljundi2016expert}}. ExpertGate is a data-based approach that learns an expert model for every task $E_1, E_2, \cdots E_n$, where every expert is adapted from the most related task. An auto-encoder model is trained for every task $AE_1, AE_2, \cdots AE_n$. These auto-encoders help determine the most related expert at test time given an example input $x$. The expert is then to make the prediction on $x$.  Note the memory {storage} requirements of ExpertGate is $n$ times the number of parameters of a single model 
which might limit its practicality. 

\noindent
\textbf{(5) Incremental Moment Matching~\cite{lee2017overcoming}} ({IMM}). 
For $N$ sequential tasks, IMM finds the optimal parameter $\mu^*_{1:N}$ and $\Sigma^*_{1:N}$ of the Gaussian approximation function $q_{1:N}$ from the posterior parameter for each $n^{th}$ task, $(\mu_n,\Sigma_n)$. At the end of the learned sequence, the obtained models are merged through a  first or second moment matching. 
Similarly to ExpertGate, IMM needs to store all models - at least if one wants to be able to add more tasks in the future.
We find the mode IMM to work consistently better than the mean IMM so we report it in our experiments.  




\noindent  \textbf{(6) Joint Training } (Joint):
In joint training, the data is not divided into tasks and the model is trained on the entire training data at once. As such, it violates the LLL assumption. This can be seen as an upper bound for all LLL methods that we evaluate.


\subsection{Adapting LLL methods to fact learning} We use the joint-embedding architecture proposed in~\cite{elhoseiny2017sherlock} as our backbone architecture to compare the evaluated   methods. We chose this architecture due its superior performance compared to other joint-embedding models like~\cite{gong2014multi,kiros2014unifying,romera2015embarrassingly} and its competitive performance to multi-class cross-entropy.
The main difference between joint embedding models and standard  classification models is in the output layer. Instead of a softmax output, the last layer in a joint-embedding model consists of a projection onto a joint embedding space. 
This allows exploiting the semantic relation between facts as well as the structure in the data, as explained before.
However, as discussed in the related work section, this is problematic for some of the LLL methods, such as~\cite{lee2017overcoming,triki2017encoder}
that assume a different output space for each task. 
This makes the problem  challenging and may raise other forgetting aspects. Note that we used the same data loss term in all the evaluated methods in the previous section. 

\vspace{-2mm}
\section{Experiments}
\label{sec:results}
We first present the evaluation metrics, then evaluate the different methods on our benchmarks and discuss the results, and finally we provide a more detailed analysis on long-tail, knowledge acquisition over time, and few-shot learning.
\subsection{Fact Learning Evaluation metrics}
\label{sec:evalmetrics}

{\bf Evaluation Metric (Standard vs Generalized). }
A central concept of LLL is that at a given time $n$ we can only observe a subset $T_n$ of the labeled training data $T_n = \{ (\mathbf{x}_k, y_k) \}_{k=1}^{M_n}\subset\mathcal{D}$ . 
Over time, we learn from different tasks $T_1,T_2,\ldots,T_N$.
The categories in the different tasks are not intersecting, i.e., if
	 $Y_n$ is the set of all category labels in task $T_n$ 
 then 
$Y_{n} \cap Y_{n'}=\emptyset, \forall n\neq n'$. Let $\mathcal{Y}$ denote the entire label space covered by all tasks, i.e., $\mathcal{Y} = \cup { Y_n}, \forall n$.
%
Many existing works assume that one does not have to disambiguate between different tasks, i.e. for a predictive function $f_y: X\mapsto \mathbb{R}$, 
we compute $A_{T_n \to {Y}_n}$ as the accuracy of classifying test data from  $T_n$  (the ${n^{th}}$ task)  into ${Y}_n$ (the label space of ${T}_n$). The accuracy is computed per task.
\begin{equation}
\vspace{-2mm}
\small
\textbf{Standard LLL (S-LLL) Accuracy: } A_{T_n \to {Y}_n}=\frac{1}{M_n}\sum_n^{M_n} 1[y_k = \arg\max_{y'\in {Y}_n}  f_{y'}(x_k) ] 
\label{ePredict1}
\end{equation}
where $y_k$ is the ground truth label for instance $x_k$.
This metric assumes that at test time one knows the task of the input image. This is how most existing works are evaluated.
However, this ignores the fact that determining the right task can be hard, especially when tasks are related. Therefore, we also evaluate across all tasks, which we refer to as \emph{Generalized LLL}.
\begin{equation}
\small
\textbf{Generalized LLL (G-LLL) Accuracy: } A_{T_n \to \mathcal{Y}}=\frac{1}{M_n}\sum_k^{M_n} 1[y_k = \arg\max_{y'\in \mathcal{Y}}  f_{y'}(x_k) ] 
\label{ePredict2}
\vspace{-2mm}
\end{equation}
In the generalized LLL metric, the search space at evaluation time covers the entire label space across tasks (i.e., $\mathcal{Y}$).  
Hence, we compute $A_{T_n \to \mathcal{Y}}$ as  the accuracy of classifying test data from  $T_n$  (the ${n^{th}}$ task) into $\mathcal{Y}$ (the entire label space) which is more realistic in many cases. 
In our experiments, $f_y(x)$  is a visual-semantic embedding model, i.e.,  $f_y(x) = s(\phi(x),\psi(y))$ where $s(\cdot, \cdot)$ is a similarity function between the visual embedding of image $x$ denoted by $\phi(x)$ and the semantic embedding of label $y$ denoted by $\psi(y)$. $\phi(x)$ is typically a CNN sub-network and $\psi(y)$ is a semantic embedding function of the label $y$ (e.g., word2vec~\cite{mikolov2013distributed}).  The above two metrics can easily be generalized to Top K standard and generalized accuracy that we use in our experiments.  

For each metric, we summarize results by averaging over tasks (``mean'') and over examples (``mean over examples''), creating slightly different results when tasks are not balanced. 
\noindent \textbf{Similarity Measure between Tasks  (word2vec, SPO overlap).} As an analysis tool, we measure similarity between tasks in both the Semantic and the Random splits using two metrics. In the first metric, the similarity is measured by the cosine similarity between average word2vec representation of the facts in each task. 
 In the second metric, we computed the overlap  between the two tasks, separately for  S, P, and O. For example, to compute the overlap in S,  we first compute the number of intersecting unique subjects and divide that by the union of unique subjects in both tasks. This results in a ratio between 0 and 1 that we compute for subjects and similarly for objects and predicates. Based on these three ratios, we compute their geometric mean 
 as an indicator for the similarity between the two tasks. We denote this measure as the SPO overlap. 
\subsection{Results}
In this section we compare several state-of-the art LLL approaches on the mid-scale and the large-scale LLL benchmark which we introduced in Sec.~\ref{sec:benchmark}. 
Tables~\ref{tab:tab:tbl_midscale_random} and~\ref{tab:tab:tbl_midscale_semantic} show the Top5 accuracy for the random and the semantic splits on the mid-scale dataset. Each table shows the performance using the standard  metric (Eq. ~\ref{ePredict1}) and the generalized metric (Eq.~\ref{ePredict2}). 
For the two large-scale benchmarks, the results are reported in Tables~\ref{tab:lsc_random_standard},~\ref{tab:lsc_random_generalized}
~\ref{tab:lsc_semantic_standard} and  ~\ref{tab:lsc_semantic_generalized}.
Note that the reported Joint Training  violates the LLL setting as it
trains on all data jointly. 
Looking at these results, we make the following observations:

%
%
%
%
%
%

\begin{table}[b!]
\vspace{-2mm}
\centering
\scalebox{0.55}
{
\begin{tabular}{|l|c|c|c|c|c|c|c|c|c|c|c|c|c|c|} \hline 
\boldmath\textbf{Random Split} & \multicolumn{6}{|c|}{\boldmath\textbf{standard metric }} & \multicolumn{6}{|c|}{\boldmath\textbf{generalized metric }} & \multicolumn{2}{|c|}{\boldmath\textbf{Drop (standard to generalized)}}    \\ \hline 
 & \multicolumn{1}{|c|}{\boldmath\textbf{T1}} & \multicolumn{1}{|l|}{\boldmath\textbf{T2}} & \multicolumn{1}{|c|}{\boldmath\textbf{T3}} & \multicolumn{1}{|c|}{\boldmath\textbf{T4}} & \multicolumn{1}{|c|}{\boldmath\textbf{mean}} & \multicolumn{1}{|c|}{\boldmath\textbf{mean over examples}} & \multicolumn{1}{|c|}{\boldmath\textbf{T1}} & \multicolumn{1}{c}{\boldmath\textbf{T2}} & \multicolumn{1}{|c|}{\boldmath\textbf{T3}} & \multicolumn{1}{|c|}{\boldmath\textbf{T4}} & \multicolumn{1}{|c|}{\boldmath\textbf{mean}} & \multicolumn{1}{|c|}{\boldmath\textbf{mean over examples}} & \multicolumn{1}{|c|}{\boldmath\textbf{over tasks}} & \multicolumn{1}{|c|}{\boldmath\textbf{over examples}}\\ \hline 
\boldmath\textbf{ExpertGate} & 79.6 & 59.25 & 62.92 & 58.75 & 65.13 & 64.88 & 53.1 & 44.83 & 37.03 & 40.66 & 43.9 & 43.69 & 21.22 & 21.18\\ \hline 
\boldmath\textbf{FineTune} & 76.41 & 46.18 & 52.44 & 88.32 & 65.84 & 66.11 & 42.06 & 22.5 & 17.84 & 83.15 & 41.39 & 42.02 & 24.45 & 24.09\\ \hline 
\boldmath\textbf{IMM} & 85.2 & 75.15 & 83.66 & 69.27 & 78.32 & 78.15 & 63.39 & 62.13 & 67.58 & 43.06 & \boldmath\textbf{59.04} & \boldmath\textbf{58.77} & 19.28 & 19.38\\ \hline 
\boldmath\textbf{Int.Synapses} & 82.31 & 65.28 & 68.64 & 87.03 & 75.81 & 75.94 & 49.37 & 39.92 & 38.45 & 74.52 & 50.57 & 50.95 & 25.25 & 24.98\\ \hline 
\boldmath\textbf{MAS} & 86.76 & 70.89 & 75.87 & 85.06 & \boldmath\textbf{79.65} & \boldmath\textbf{79.68} & 55.23 & 48.62 & 51.1 & 71.72 & {56.67} & {56.94} & 22.98 & 22.74\\ \hline  \hline 
\boldmath\textbf{Joint} & 88.66 & 78.38 & 87.82 & 75.91 & 82.69 & 82.57 & 75.81 & 68.45 & 79.03 & 60.82 & 71.03 & 70.87 & 11.67 & 11.7 \\ \hline \end{tabular}
}
\caption{Mid-scale Dataset (Random Split) Top 5 Accuracy} 
\label{tab:tab:tbl_midscale_random}
\vspace{-4mm}
\end{table}
\begin{table}[b!]
\centering
\scalebox{0.55}
{
\begin{tabular}{@{}|l|c|c|c|c|c|c|c|c|c|c|c|c|c|c|@{}} \hline
\boldmath\textbf{Semantic Split} &   \multicolumn{6}{|c|}{\boldmath\textbf{standard}}  & \multicolumn{6}{|c|}{\boldmath\textbf{generalized}}    & \multicolumn{2}{|c|}{\boldmath\textbf{Drop (standard to generalized)}}  \\ \hline
 & \multicolumn{1}{l}{\boldmath\textbf{T1}} & \multicolumn{1}{l}{\boldmath\textbf{T2}} & \multicolumn{1}{c}{\boldmath\textbf{T3}} & \multicolumn{1}{|c|}{\boldmath\textbf{T4}} & \multicolumn{1}{|c|}{\boldmath\textbf{mean}} & \multicolumn{1}{|c|}{\boldmath\textbf{mean over examples}} & \multicolumn{1}{|c|}{\boldmath\textbf{T1}} & \multicolumn{1}{|c|}{\boldmath\textbf{T2}} & \multicolumn{1}{|c|}{\boldmath\textbf{T3}} & \multicolumn{1}{|c|}{\boldmath\textbf{T4}} & \multicolumn{1}{|c|}{\boldmath\textbf{mean}} & \multicolumn{1}{|c|}{\boldmath\textbf{mean over examples}} & \multicolumn{1}{|c|}{\boldmath\textbf{over tasks}} & \multicolumn{1}{|c|}{\boldmath\textbf{over examples}}\\ \hline
\boldmath\textbf{ExpertGate} & 62.11 & 62.44 & 59.4 & 12.5 & 49.11 & 59.39 & 55.57 & 50.87 & 49.61 & 9.49 & {\bf 41.38} & {\bf 51.83} & 7.73 & 7.57\\ \hline
\boldmath\textbf{FineTune} & 16.24 & 35.63 & 31.71 & 15.19 & 24.69 & 21.3 & 8.25 & 0 & 0 & 15.19 & 5.86 & 6.07 & 18.83 & 15.24\\ \hline
\boldmath\textbf{IMM} & 64.26 & 87.52 & 63.27 & 12.82 & {\bf 56.97} & {\bf 64.33} & 38.75 & 30.16 & 43.28 & 8.70 & 30.22 & 37.31 & 26.74 & 27.02\\ \hline
\boldmath\textbf{Int.Synapses} & 16.48 & 35.69 & 32.01 & 8.54 & 23.18 & 21.23 & 8.25 & 0 & 0 & 8.54 & 4.2 & 5.77 & 18.98 & 15.46\\ \hline
\boldmath\textbf{MAS} & 28.19 & 47.91 & 34.8 & 12.97 & 30.97 & 30.96 & 8.34 & 0.13 & 0 & 12.97 & 5.36 & 6.04 & 25.61 & 24.92\\ \hline \hline 
\boldmath\textbf{Joint} & 80.14 & 53.12 & 81.47 & 21.2 & 58.98 & 74.75 & 77.34 & 39.55 & 79.2 & 18.99 & 53.77 & 70.87 & 5.22 & 3.87\\ \hline
\end{tabular}
\vspace{-4mm}
}
\caption{Mid-scale Dataset (Semantic Split)} 
\label{tab:tab:tbl_midscale_semantic}
\end{table}

\begin{table}[h!]
\centering
\vspace{-4mm}
\scalebox{0.8}
{
\begin{tabular}{@{}|l|c|c|c|c|c|c|c|c|c|c|@{}} \hline 
\boldmath\textbf{Random } & \multicolumn{1}{|c|}{T1} & \multicolumn{1}{|c|}{T2} & \multicolumn{1}{c}{T3} & \multicolumn{1}{|c|}{T4} & \multicolumn{1}{c}{T5} & \multicolumn{1}{|c|}{T6} & \multicolumn{1}{c}{T7} & \multicolumn{1}{|c|}{T8} & \multicolumn{1}{c}{mean} & \multicolumn{1}{|c|}{mean over examples}\\ \hline 
\boldmath\textbf{ExpertGate} & 16.37 & 20.49 & 28.36 & 22.2 & 37.75 & 12.52 & 14.14 & 24.37 & 22.02 & 20.95\\ \hline 
\boldmath\textbf{Finetune} & 15.53 & 23.56 & 23.43 & 19.22 & 23.44 & 19.53 & 23.81 & 66.13 & 26.83 & 26.59\\ \hline 
\boldmath\textbf{IMM} & 24.57 & 30.72 & 32.14 & 27.89 & 37.23 & 25 & 20.65 & 26.41 & 28.08 & 27.53\\ \hline 
\boldmath\textbf{Int.Synapses} & 18.28 & 27.23 & 27.11 & 23.3 & 28.85 & 23.49 & 25.76 & 53.43 & 28.43 & 28.06\\ \hline 
\boldmath\textbf{MAS} & 21.32 & 33.32 & 32.82 & 28.58 & 34.93 & 27.16 & 29.71 & 52.22 & {\bf 32.5} & {\bf 32{.00}}\\ \hline 
\end{tabular}
}
\caption{Large Scale Random Split (Standard Performance) Top 5 Accuracy} 
\label{tab:lsc_random_standard}
\vspace{-2mm}
\end{table}

\begin{table}[h!]
\vspace{-4mm}
\centering
\scalebox{0.8}
{
\begin{tabular}{@{}|l|c|c|c|c|c|c|c|c|c|c|@{}} \hline 
\boldmath\textbf{Random } & \multicolumn{1}{|c|}{T1} & \multicolumn{1}{|c|}{T2} & \multicolumn{1}{c}{T3} & \multicolumn{1}{|c|}{T4} & \multicolumn{1}{c}{T5} & \multicolumn{1}{|c|}{T6} & \multicolumn{1}{c}{T7} & \multicolumn{1}{|c|}{T8} & \multicolumn{1}{c}{mean} & \multicolumn{1}{|c|}{mean over examples}\\ \hline 
\boldmath\textbf{ExpertGate} & 12.99 & 20.77 & 25.19 & 17.72 & 35.17 & 9.62 & 11.64 & 21.75 & 19.36 & 15.34\\ \hline
\boldmath\textbf{Finetune} & 12.18 & 21.38 & 19.98 & 15.68 & 19.85 & 16.11 & 17.48 & 59.29 & 22.74 & 18.93\\ \hline
\boldmath\textbf{IMM} & 21.21 & 29.02 & 30.5 & 25.38 & 34.01 & 23.42 & 18.07 & 24.26 & 25.73 & 20.91\\ \hline
\boldmath\textbf{Int.Synapses} & 13.79 & 24.99 & 23.58 & 19.01 & 26.4 & 21.56 & 20.95 & 47.69 & 24.75 & 19.92\\ \hline
\boldmath\textbf{MAS} & 16.13 & 29.52 & 28.28 & 23.1 & 30.28 & 24.5 & 24.34 & 47.21 & {\bf 27.92} & {\bf 22.48}\\  \hline 
\end{tabular}
}
\caption{Large Scale Random Split (Generalized Performance) Top 5 Accuracy}
\label{tab:lsc_random_generalized}
\vspace{-2mm}
\end{table}

\noindent \textbf{(1) } The generalized LLL accuracy  is always significantly lower than the standard LLL accuracy. On the large scale benchmarks it is  on average several percent lower: $7.99\%$ and $11.59\%$ for the random and the semantic splits, respectively. While the large-scale benchmarks are more challenging than the mid-scale benchmarks, as apparent from the reported accuracies, the drop in performance when switching to the generalized accuracy on the mid-scale benchmarks is significantly larger: $20.59 \%$ and $18.16 \%$, respectively.
This could be due to more overlap between tasks on the large-scale dataset as we discuss later, which reduces forgetting leading to better discrimination across tasks. 


\noindent \textbf{(2) } The LLL performance of the random split is much better compared to the semantic split. Note that the union of the test examples across tasks on both splits are the same. Hence, the `mean over examples'' performance on the random and semantic splits are comparable.  Looking at the performance of the evaluated methods on both random and semantic splits on the large scale dataset, the average relative gain in performance over the methods by using the random split instead of the semantic split is $61.74\%$ for the generalized metrics. This gain is not observed for ExpertGate which has only $2.77\%$ relative gain when moving to the random split (small compared to other methods).  We discuss ExpertGate behavior in a separate point below. The same ratio goes up to $569.03\%$  on the mid-scale dataset excluding ExpertGate. What explains these results is that the similarity between tasks in the random split is much higher in the large-scale dataset  compared to the mid-scale dataset (i.e., 0.96  vs  0.22 using the word2vec metric and 0.84 vs 0.25 using the SPO metric -- see Table~\ref{tab:lsc_task_correlation} for the task correlation in the LSc dataset and the corresponding table for the mid-scale dataset in the supplementary. This shows the learning difficulty of the semantic split and partially explains the poor performance.

\begin{table}[t!]
\vspace{-4mm}
\centering
\scalebox{0.8}
{
\begin{tabular}{@{}|l|c|c|c|c|c|c|c|c|c|c|@{}} \hline 
\boldmath\textbf{Semantic } & \multicolumn{1}{|c|}{T1} & \multicolumn{1}{|c|}{T2} & \multicolumn{1}{c}{T3} & \multicolumn{1}{|c|}{T4} & \multicolumn{1}{c}{T5} & \multicolumn{1}{|c|}{T6} & \multicolumn{1}{c}{T7} & \multicolumn{1}{|c|}{T8} & \multicolumn{1}{c}{mean} & \multicolumn{1}{|c|}{mean over examples}\\ \hline 
\boldmath\textbf{ExpertGate} & 6.97 & 11.01 & 35.6 & 34.61 & 14.58 & 21.32 & 16.36 & 13.28 & {\bf 19.22} & \boldmath\textbf{20.15}\\ \hline 
\boldmath\textbf{Finetune} & 5.55 & 11.17 & 13.65 & 24.04 & 10.84 & 12.68 & 19.41 & 39.41 & 17.09 & 17.91\\ \hline 
\boldmath\textbf{IMM} & 9.49 & 9.25 & 16.90 & 30.95 & 11.05 & 33.92 & 18.2 & 10.99 & 17.59 & 14.81\\ \hline 
\boldmath\textbf{Int.Synapses} & 5.47 & 13.3 & 14.95 & 25.23 & 12.43 & 14.4 & 20.18 & 29.8 & 16.97 & 17.49\\ \hline 
\boldmath\textbf{MAS} & 6.36 & 14.16 & 19.51 & 26.25 & 13.25 & 15.22 & 20.57 & 28.59 & 17.99 & 18.75\\ \hline \hline
\boldmath\textbf{Joint} & 11.62& 5.90 &36.26 &37.56& 28.16 &16.16 & 14.32 &12.85 & 20.35 & 23.41\\ \hline
\end{tabular}
}
\caption{Large Scale Semantic Split (Standard Performance) Top 5 Accuracy} 
\label{tab:lsc_semantic_standard}
\vspace{-2mm}
\end{table}

\begin{table}[t!]
\vspace{-4mm}
\centering
\scalebox{0.8}
{
\begin{tabular}{@{}|l|c|c|c|c|c|c|c|c|c|c|@{}}  \hline 
\boldmath\textbf{Semantic } & \multicolumn{1}{|c|}{T1} & \multicolumn{1}{|c|}{T2} & \multicolumn{1}{c}{T3} & \multicolumn{1}{|c|}{T4} & \multicolumn{1}{c}{T5} & \multicolumn{1}{|c|}{T6} & \multicolumn{1}{c}{T7} & \multicolumn{1}{|c|}{T8} & \multicolumn{1}{c}{mean} & \multicolumn{1}{|c|}{mean over examples}\\ \hline 
\boldmath\textbf{ExpertGate} & 5.18 & 7.62 & 35.33 & 20.35 & 8.99 & 16.59 & 6.21 & 7.19 & {\bf 13.43} & {\bf 14.91}\\ \hline 
\boldmath\textbf{Finetune} & 1.58 & 8.56 & 0.07 & 2.06 & 5.88 & 2.86 & 4.77 & 37.9 & 7.96 & 9.75\\ \hline 
\boldmath\textbf{IMM} & 8.34 & 5.06 & 0.18 & 13.27 & 0.52 & 21.48 & 11.21 & 3.26 & 7.91 & 4.15\\ \hline 
\boldmath\textbf{Int.Synapses} & 1.71 & 10.82 & 0.22 & 2.84 & 5.87 & 4.77 & 6.36 & 28.26 & 7.61 & 8.70\\ \hline 
\boldmath\textbf{MAS} & 1.79 & 11.35 & 0.64 & 4.25 & 4.76 & 5.36 & 6.2 & 27.35 & 7.71 & 8.54\\ \hline \hline
\boldmath\textbf{Joint} & 10.20 &	4.86 &	37.71 &	33.52 &	25.09	& 3.17 &	4.83 &	8.43 & 15.98& 20.68\\ \hline
\end{tabular}
}
\caption{Large Scale Semantic Split (Generalized Performance) Top 5 Accuracy} 
\label{tab:lsc_semantic_generalized}
\vspace{-2mm}
\end{table}

\begin{table}[t!]
\vspace{-4mm}
\centering
\scalebox{0.65}
{
\begin{tabular}{@{}|l|c|c|c|c|c|c|c|c|l|c|c|c|c|c|c|c|c|@{}} \hline
\multicolumn{9}{|l|}{\boldmath\textbf{Semantic(0.07 mean similarity)} } & \multicolumn{9}{|l|}{\boldmath\textbf{Random (0.96 mean similarity)}} \\  \hline x
 & \multicolumn{1}{l}{\boldmath\textbf{T1}} & \multicolumn{1}{|l|}{\boldmath\textbf{T2}} & \multicolumn{1}{l}{\boldmath\textbf{T3}} & \multicolumn{1}{|l|}{\boldmath\textbf{T4}} & \multicolumn{1}{|l|}{\boldmath\textbf{T5}} & \multicolumn{1}{|l|}{\boldmath\textbf{T6}} & \multicolumn{1}{|l|}{\boldmath\textbf{T7}} & \multicolumn{1}{|l|}{\boldmath\textbf{T8}} &  & \multicolumn{1}{|l|}{\boldmath\textbf{T1}} & \multicolumn{1}{|l|}{\boldmath\textbf{T2}} & \multicolumn{1}{|l|}{\boldmath\textbf{T3}} & \multicolumn{1}{l}{\boldmath\textbf{T4}} & \multicolumn{1}{|l|}{\boldmath\textbf{T5}} & \multicolumn{1}{|l|}{\boldmath\textbf{T6}} & \multicolumn{1}{l}{\boldmath\textbf{T7}} & \multicolumn{1}{|l|}{\boldmath\textbf{T8}}\\ \hline
\boldmath\textbf{T1} &1& 0.32 & -0.28 & -0.18 & -0.45 & 0.23 & 0.16 & 0.15 & \boldmath\textbf{T1} &1& 0.97 & 0.97 & 0.97 & 0.95 & 0.97 & 0.97 & 0.97\\ \hline
\boldmath\textbf{T2} & 0.32 &1& -0.37 & -0.11 & -0.68 & 0.21 & 0.36 & 0.29 & \boldmath\textbf{T2} & 0.97 &1& 0.97 & 0.95 & 0.95 & 0.96 & 0.96 & 0.96\\ \hline
\boldmath\textbf{T3} & -0.28 & -0.37 &1& 0.25 & 0.05 & -0.26 & -0.22 & -0.52 & \boldmath\textbf{T3} & 0.97 & 0.97 &1& 0.97 & 0.95 & 0.96 & 0.97 & 0.96\\ \hline
\boldmath\textbf{T4} & -0.18 & -0.11 & 0.25 &1& -0.08 & -0.01 & -0.12 & -0.41 & \boldmath\textbf{T4} & 0.97 & 0.95 & 0.97 &1& 0.95 & 0.95 & 0.97 & 0.96\\ \hline
\boldmath\textbf{T5} & -0.45 & -0.68 & 0.05 & -0.08 &1& -0.26 & -0.36 & -0.04 & \boldmath\textbf{T5} & 0.95 & 0.95 & 0.95 & 0.95 &1& 0.95 & 0.96 & 0.93\\\hline
\boldmath\textbf{T6} & 0.23 & 0.21 & -0.26 & -0.01 & -0.26 &1& 0.23 & 0.26 & T6 & 0.97 & 0.96 & 0.96 & 0.95 & 0.95 &1& 0.96 & 0.96\\ \hline
\boldmath\textbf{T7} & 0.16 & 0.36 & -0.22 & -0.12 & -0.36 & 0.23 &1& 0.35 & \boldmath\textbf{T7} & 0.97 & 0.96 & 0.97 & 0.97 & 0.96 & 0.96 &1& 0.95\\ \hline
\boldmath\textbf{T8} & 0.15 & 0.29 & -0.52 & -0.41 & -0.04 & 0.26 & 0.35 &1& \boldmath\textbf{T8} & 0.97 & 0.96 & 0.96 & 0.96 & 0.93 & 0.96 & 0.95 & 1\\ \hline \hline
  \multicolumn{9}{|l|}{\boldmath\textbf{Semantic (0.238  g-mean of S,P, and O overlap)}}  &  \multicolumn{9}{|l|}{\boldmath\textbf{Random (0.453 g-mean of S,P, O overlap)}}   \\ \hline
\boldmath\textbf{T1} & 1 & 0.09 & 0.08 & 0.11 & 0.05 & 0.06 & 0.1 & 0.08 & \boldmath\textbf{T1} & 1 & 0.39 & 0.38 & 0.38 & 0.35 & 0.39 & 0.4 & 0.38\\ \hline
\boldmath\textbf{T2} & 0.09 & 1 & 0.12 & 0.15 & 0.15 & 0.08 & 0.27 & 0.28 & \boldmath\textbf{T2} & 0.39 & 1 & 0.38 & 0.37 & 0.35 & 0.37 & 0.38 & 0.38\\ \hline
\boldmath\textbf{T3} & 0.08 & 0.12 & 1 & 0.23 & 0.2 & 0.04 & 0.12 & 0.1 & \boldmath\textbf{T3} & 0.38 & 0.38 & 1 & 0.38 & 0.36 & 0.38 & 0.4 & 0.38\\ \hline
\boldmath\textbf{T4} & 0.11 & 0.15 & 0.23 & 1 & 0.15 & 0.08 & 0.14 & 0.12 & \boldmath\textbf{T4} & 0.38 & 0.37 & 0.38 & 1 & 0.37 & 0.37 & 0.37 & 0.37\\ \hline
\boldmath\textbf{T5} & 0.05 & 0.15 & 0.2 & 0.15 & 1 & 0.05 & 0.12 & 0.18 & \boldmath\textbf{T5} & 0.35 & 0.35 & 0.36 & 0.37 & 1 & 0.36 & 0.37 & 0.36\\ \hline
\boldmath\textbf{T6} & 0.06 & 0.08 & 0.04 & 0.08 & 0.05 & 1 & 0.12 & 0.1 & \boldmath\textbf{T6} & 0.39 & 0.37 & 0.38 & 0.37 & 0.36 & 1 & 0.38 & 0.38\\ \hline
\boldmath\textbf{T7} & 0.1 & 0.27 & 0.12 & 0.14 & 0.12 & 0.12 & 1 & 0.28 & \boldmath\textbf{T7} & 0.4 & 0.38 & 0.4 & 0.37 & 0.37 & 0.38 & 1 & 0.38\\ \hline
\boldmath\textbf{T8} & 0.08 & 0.28 & 0.1 & 0.12 & 0.18 & 0.1 & 0.28 & 1 & \boldmath\textbf{T8} & 0.38 & 0.38 & 0.38 & 0.37 & 0.36 & 0.38 & 0.38 & 1\\ \hline

\end{tabular}
\vspace{-3mm}
}
\caption{Large Scale Task Similarities using average Word2vec space (top-part) and geometric mean S,P, and O overlap  (bottom-part)} 
\label{tab:lsc_task_correlation}
\vspace{-8mm}
\end{table}

\noindent \textbf{(3) } ExpertGate is the best performing model on the semantic split. However, it is among the worst performing models on the random split. We argue that this is due to the setup of the semantic split, where sharing across tasks is minimized. This makes each task model behave like an expert of a restricted concept space, which follows the underlying assumption of how ExpertGate works. However, 
this advantage comes at the  expense of storing one model for every task which can be  expensive w.r.t. storage requirements which might not always be feasible 
as the number of tasks increases. 
Additionally, having separate models, requires to select a model at test time and also removes the ability to benefit from knowledge learnt with later tasks, in case there is a semantic overlap between tasks. This can be seen on the random split on the mid-scale dataset (see Table~\ref{tab:tab:tbl_midscale_random}) where ExpertGate underperforms several other LLL models:
43.69\% generalized accuracy for ExpertGate vs 58.77\% generalized accuracy for the best performing model. Similarly on the large scale dataset, ExpertGate performs significantly lower for the random split (15.34\% generalized accuracy for ExpertGate vs  22.48\% generalized accuracy for the best performing model); see Table~\ref{tab:lsc_semantic_generalized}. The shared information across tasks on the random split is high which violates the assumption of expert selection in the ExpertGate method and hence explains its relatively  poor performance on the random split. 

\noindent \textbf{(4) } For the \textit{midscale dataset} and with the generalized metric, Incremental Moment Matching (IMM) is the best performing of the  model-based methods using a single model (Finetune, IMM, Int.Synapses, MAS) on both the random and the semantic splits (see Tables \ref{tab:tab:tbl_midscale_random},\ref{tab:tab:tbl_midscale_semantic}). 
Only for the random split evaluated with the standard metric MAS 
is slightly better, 
indicating that MAS might be better at the task level. 
We hypothesize that IMM benefits from its access to the distribution of the parameters after training each task before the distributions' mode is computed. This is an advantage that MAS and Int.Synapses do not have and hence the IMM model can generalize better across tasks. For the \textit{large-scale dataset}, we observe that MAS is performing better  than IMM on both the random and the semantic split, but especially on the random split; see Table~\ref{tab:lsc_random_generalized}. This may be because MAS has a better capability to learn low-shot classes  as we discuss later in our \textit{Few-shot Analysis}; see tables~\ref{tab:lsc_semantic_few_shot} and ~\ref{tab:lsc_random_few_shot}. This is due to the high similarity between the tasks as we go to that much larger scale; see Table~\ref{tab:lsc_task_correlation}. This makes the distribution of parameters that work well across tasks similar to each other and hence IMM no longer has the aforementioned advantage. 

\begin{figure*}[t]
  \centering
\vspace{-0.4cm}\hspace{-0.6cm}\includegraphics[width=0.37\textwidth]{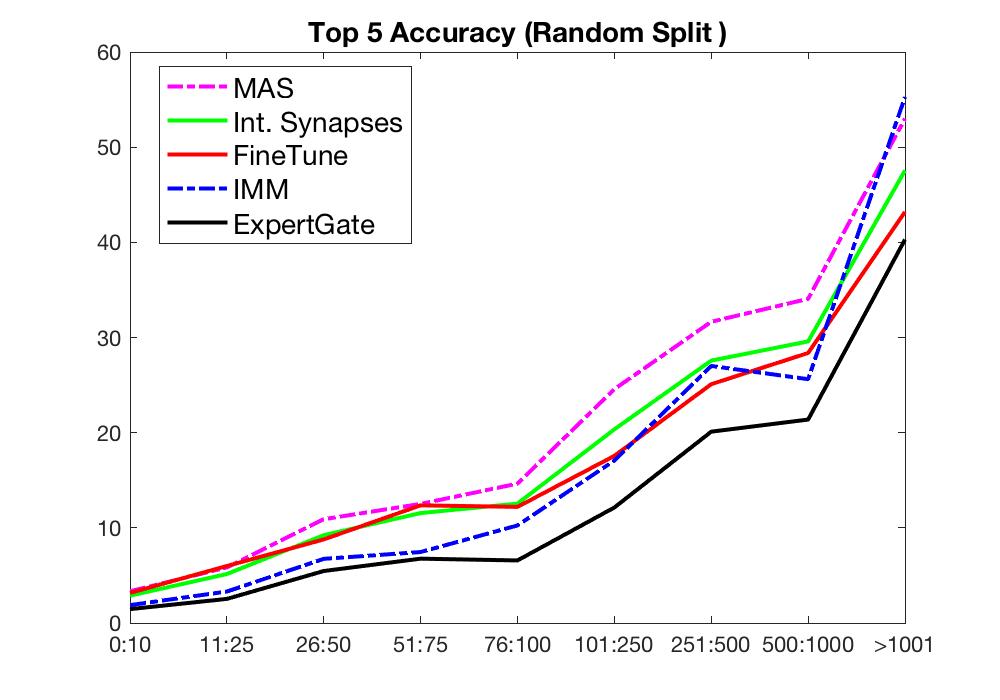}
     \hspace{-0.5cm}\includegraphics[width=0.37\textwidth]{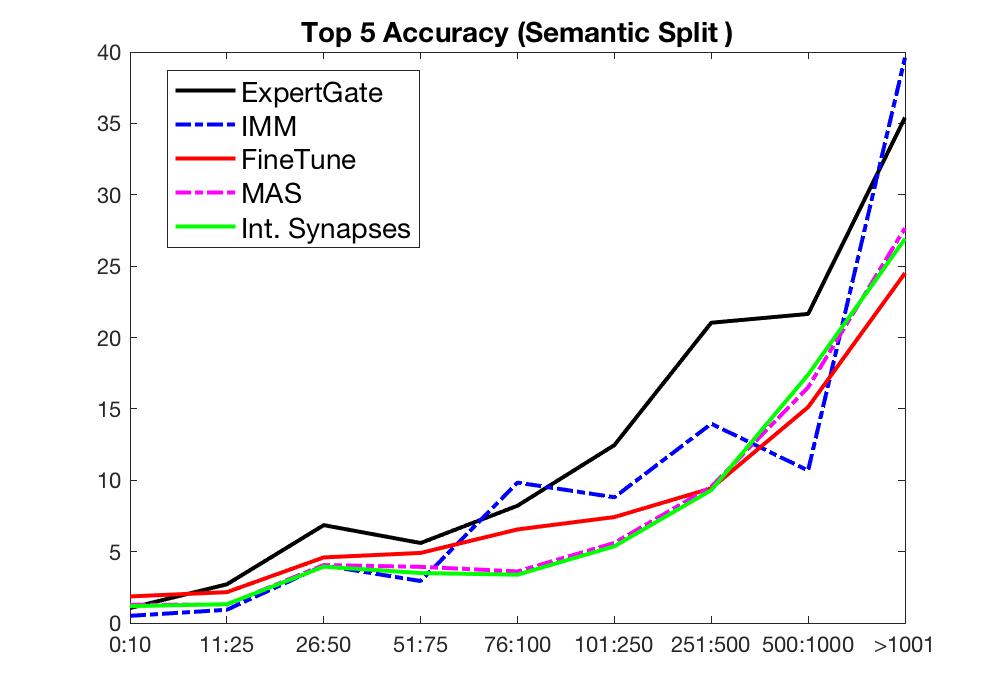}
         \hspace{-0.5cm}\includegraphics[width=0.37\textwidth]{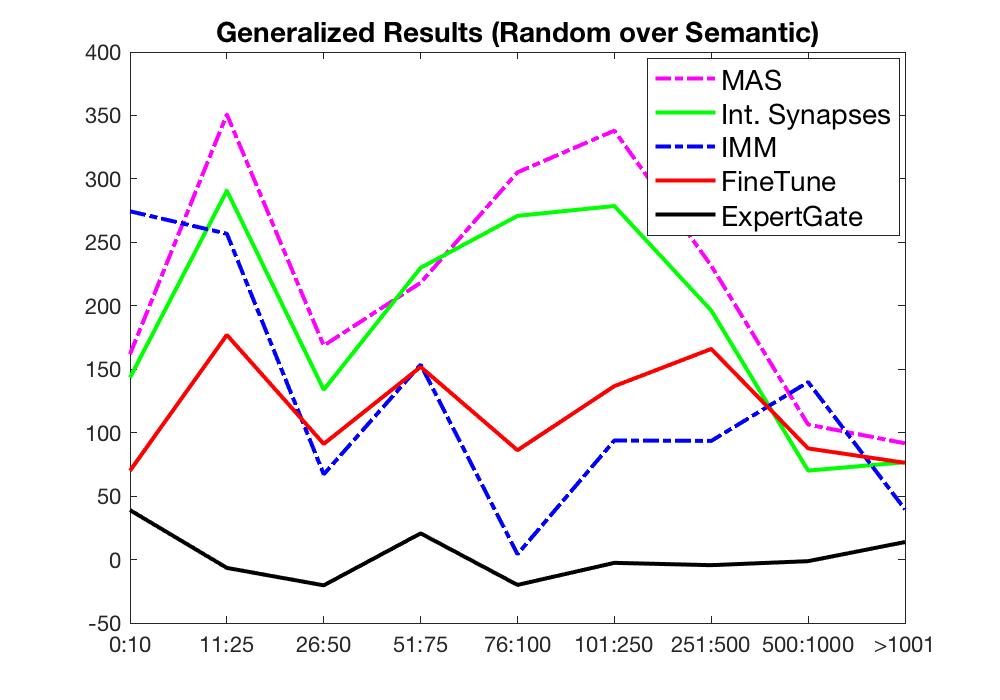}\hspace{-0.5cm}
         \vspace{-0.5cm}
      \caption{ LLFL Benchmark Long-tail analysis (Generalized Results). The x-axis in this figure shows the range of examples seen during training.  On the left and middle: the y-axis shows the generalized Top 5 Accuracy  for the random and the semantic splits. On the right: The y-axis shows the Random split improvement over the random split for each range.
      }
      \label{fig_semantic_vs_random_gen}
      \vspace{-4mm}
\end{figure*}

\subsection{Detailed Analysis}
\label{sec:detailedAnalysis}
\text{\bf Long-tail Analysis.}  We show in Fig~\ref{fig_semantic_vs_random_gen} on left and middle the head-to-tail performance on the random split and the semantic split respectively. Specifically, the figure shows the Top5 generalized accuracy over different ranges of seen examples per class (i.e., the x-axis in the figure).
On the right, the figure shows the relative improvement of the model trained on the random split over the semantic split. Using the standard metrics, the head classes perform better using  models trained on the semantic split compared to the random split. It also shows that the random splits benefit the tail-classes the most; shown in supplementary materials (Section 4). However as shown on Fig~\ref{fig_semantic_vs_random_gen} (right), random split benefits everywhere with no clear relation to the class frequency (x-axis).   

\begin{figure*}[t!]
  \centering
   \includegraphics[width=0.38\textwidth]{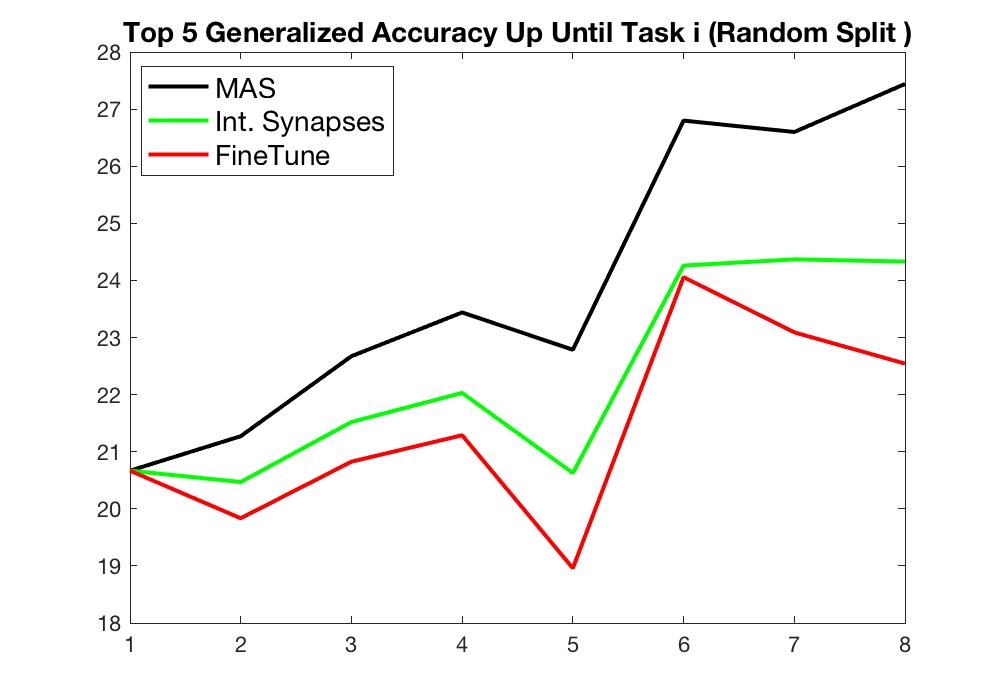}
   \includegraphics[width=0.38\textwidth] {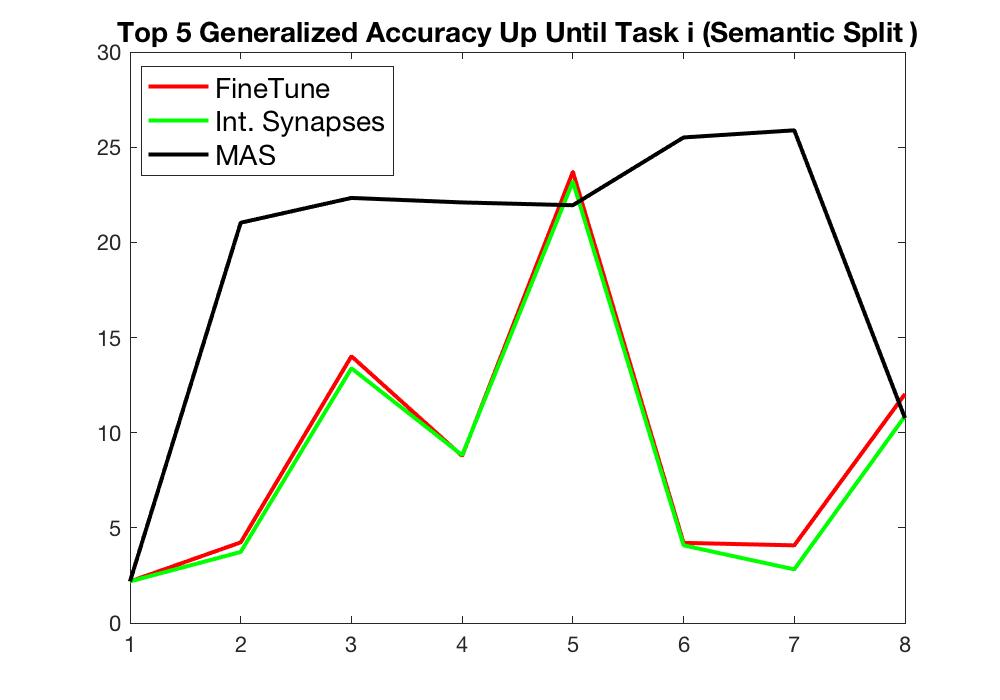}
     \vspace{-4mm}
      \caption{Gained Visual Knowledge: The x-axis shows the task number i. The y-axis shows the Top5  generalized accuracy over the entire test set up until training task i, for the random (left) and semantic (right) split respectively. 
      }
        \vspace{-3mm}
      \label{fig_semantic_radom_visual_knowledge}
\end{figure*}

\noindent\textbf{Gained Knowledge Over Time.}  Figure~\ref{fig_semantic_radom_visual_knowledge} shows the gained knowledge over time measured  by the generalized Top5 Accuracy of the entire test set of all tasks after training each task. Figure~\ref{fig_semantic_radom_visual_knowledge} (left) shows that the LLL methods tend to gain more knowledge over time when the random split is used. This is due to the high similarity between tasks which makes the forgetting over time less catastrophic. Figure~\ref{fig_semantic_radom_visual_knowledge} (right) shows that the models have difficulty gaining knowledge over time when the semantic split is used.  This is due to the low similarity between tasks which makes the forgetting over time more catastrophic. Note that the y-axis in Figure~\ref{fig_semantic_radom_visual_knowledge}  left  and right parts are comparable since it measure the performance of the entire test set which is the same on both the semantic and the random splits. 



For a principled evaluation, we consider measuring the forward and the backward transfer as defined in~\cite{lopez2017gradient}.  After each model finishes learning about the task $T_n$, we evaluate
its \emph{test} performance on all $N$ tasks.
By doing so, we construct the
matrix $R \in \mathbb{R}^{N \times N}$, where $R_{j,n}$ is the test
classification accuracy of the model on task $T_j$ after observing the last
sample from task $T_n$.  Letting $\bar{b}$ be the vector of test accuracies for
each task at random initialization, we can define the backward and the forward transfer as: $\text{ {\bf Backward Transfer: } \normalsize BWT} = \frac{1}{N-1}
\sum_{n=1}^{N-1} R_{n,N} - R_{n,n}$  and $\text{ {\bf Forward Transfer: } \normalsize FWT} = \frac{1}{N-1}
\sum_{n=2}^{N} R_{n,n-1} - \bar{b}_n$.
The larger these metrics, the better the model. If two
models have similar accuracy, the most preferable one is the one with larger BWT and
FWT. 
We used 
the generalized accuracy for computing BWT and FWT.

\begin{figure*}[t!]
  \centering
   \includegraphics[width=0.9\textwidth]{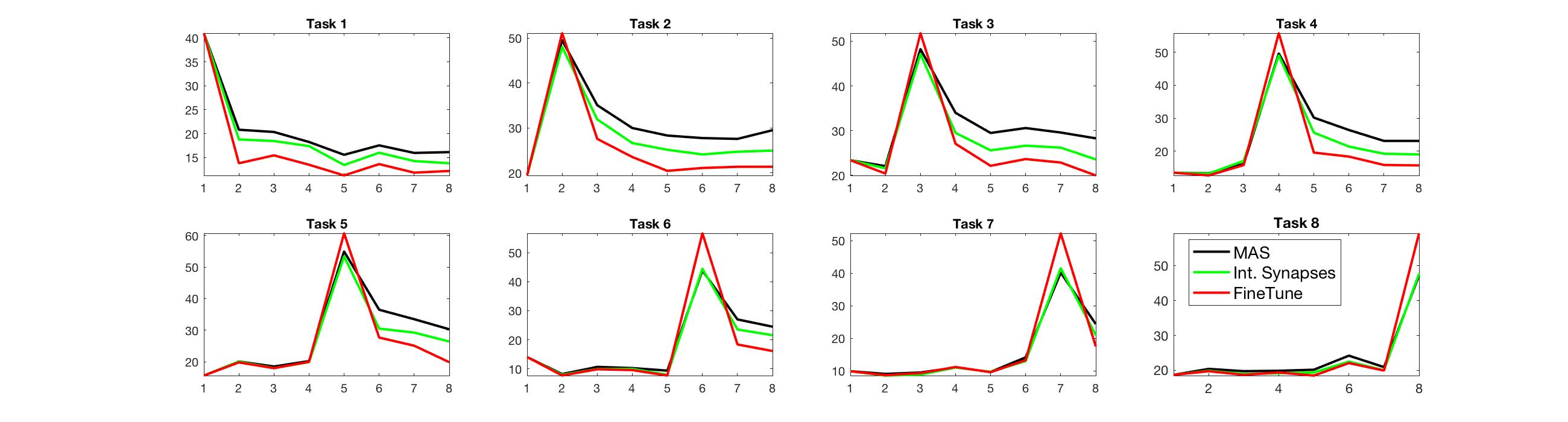}
     \vspace{-3mm}
      \caption{Gained Visual Knowledge broken down for each task on the random split: The x-axis in each sub figure shows the task number i. The y-axis shows the Random Split  Top5  generalized accuracy of the shown task after each task is learnt.}
      \label{fig_random_visual_knowledge}
        \vspace{-3mm}
\end{figure*}

\begin{figure*}[t!]
  \vspace{-2mm}
  \centering
  \includegraphics[width=0.9\textwidth]{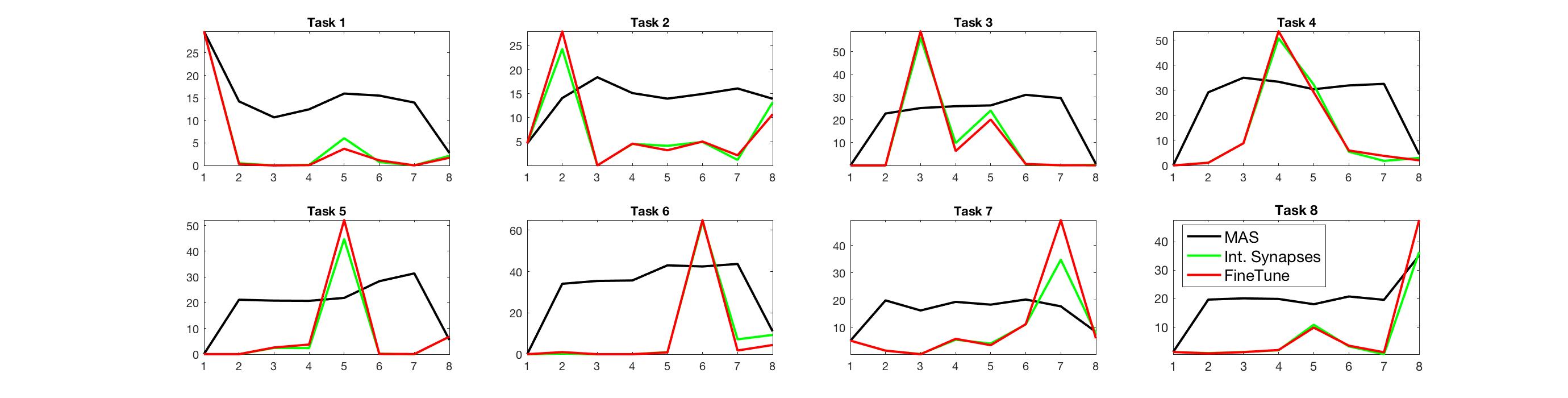}
  \vspace{-5mm}
      \caption{Gained Visual Knowledge broken down for each task on the semantic split: The x-axis in each sub figure shows the task number i. The y-axis shows the Random Split  Top5  generalized accuracy of the shown task after each task is learnt.}
      \label{fig_semantic_visual_knowledge}
        \vspace{-1mm}
\end{figure*}

\begin{wraptable}{r}{0.5\textwidth}
\vspace{-3mm}
\scalebox{0.8}
{
\begin{tabular}{@{}|l|r|r|r|@{}}  \hline 
\boldmath\textbf{Semantic } & \multicolumn{1}{|l|}{MAS} & \multicolumn{1}{|l|}{Int.Synapses} & \multicolumn{1}{|l|}{Finetune} \\ \hline
{Forward Transfer} & 0.24  &  0.08  &  0.06 \\ \hline 
{Backward Transfer} & -0.20 &  -0.37 &   -0.44  \\ \hline  \hline 
\boldmath\textbf{Random } & \multicolumn{1}{|l|}{MAS} & \multicolumn{1}{|l|}{Int.Synapses} & \multicolumn{1}{|l|}{Finetune} \\ \hline
{Forward Transfer} & 0.30  &  0.26 &   0.22 \\ \hline 
{Backward Transfer} & -0.22 &   -0.25  &  -0.35  \\ \hline  
\end{tabular}
}
\vspace{-2mm}
\caption{Large Scale Benchmark Forward and Backward Transfer for continual learning methods}
\label{tbl_fwd_bwd}
\vspace{-8mm}\end{wraptable}Figures~\ref{fig_random_visual_knowledge} and ~\ref{fig_semantic_visual_knowledge} show the performance of each task test set after training each task (from first task to last task). As expected the performance on the task $n$ set peaks after training task $n$ and the performance degrades after training subsequent tasks. Int.Synapses and Finetune show the best performance of training the current task at the expense of more forgetting on previous tasks compared to MAS. Comparing the performance of  task $n$ at the $n^{th}$ task training to its performance after training the last task as a measure of forgetting,  we can observe a lower drop on the performance on the random split compared to the semantic split; see the figures. This is also demonstrated by higher backward transfer on the random split; see Table~\ref{tbl_fwd_bwd}.


\noindent\textbf{Few-shot Analysis.} Now, we focus on analyzing the subset of the testing examples belonging to facet with few training examples. Tables ~\ref{tab:lsc_semantic_few_shot} and ~\ref{tab:lsc_random_few_shot} show  few-shot results 
on the semantic and the random split, respectively.   
As already observed earlier, the performance on the random splits is better compared to the semantic splits. We can observe here that  finetuning is the best performing approach on  average for few-shot performance on both splits. Looking closely at the results, it is not hard to see that the main gain of  finetuning is due to its high accuracy on the last task. This shows that existing LLL methods do not learn the tail and there is need to devise new methods that have a capability to learn the tail distribution in a LLL setting. 

\begin{table}[t!]
\parbox{.49\linewidth}{
\centering
\scalebox{0.72}
{
\begin{tabular}{@{}|l|r|r|r|r|r|r|r|r|r|@{}} \hline
 & \multicolumn{1}{|l|}{T1} & \multicolumn{1}{|l|}{T2} & \multicolumn{1}{|l|}{T3} & \multicolumn{1}{|l|}{T4} & \multicolumn{1}{|l|}{T5} & \multicolumn{1}{|l|}{T6} & \multicolumn{1}{|l|}{T7} & \multicolumn{1}{|l|}{T8} & \multicolumn{1}{|l|}{average}\\ \hline
\boldmath\textbf{ExpertGate} & 0.33 & 0.88 & 0.57 & 3.27 & 0.69 & 2.04 & 0 & 0.62 & 1.05\\ \hline
\boldmath\textbf{Finetune} & 0 & 0.65 & 0 & 0 & 0.3 & 2.04 & 1.76 & 10.14 & 1.86\\ \hline
\boldmath\textbf{IMM} & 1.3 & 0.03 & 0 & 0 & 0.04 & 2.65 & 0 & 0 & 0.5\\ \hline
\boldmath\textbf{Int.Synapses} & 0.11 & 0.37 & 0 & 0 & 0.28 & 2.65 & 1.14 & 4.91 & 1.18\\ \hline
\boldmath\textbf{MAS} & 0.11 & 0.31 & 0 & 0.05 & 0.22 & 3.67 & 1.24 & 4.58 & 1.27\\ \hline \hline
\boldmath\textbf{Joint} & 5.55 & 2.65 & 4.79 & 6.27 & 2.85 & 10.9 & 8.88 & 3.7 & 5.7\\ \hline
\end{tabular}
}
\caption{Few-shot ($\leq 10$) generalized Top 5 accuracy, Large-scale, Semantic Split} 
\label{tab:lsc_semantic_few_shot}
}
\hfill
\parbox{.49\linewidth}{
\centering
\scalebox{0.72}
{
\begin{tabular}{@{}|l|r|r|r|r|r|r|r|r|r|@{}} \hline
 & \multicolumn{1}{|l|}{T1} & \multicolumn{1}{|l|}{T2} & \multicolumn{1}{|l|}{T3} & \multicolumn{1}{|l|}{T4} & \multicolumn{1}{|l|}{T5} & \multicolumn{1}{|l|}{T6} & \multicolumn{1}{|l|}{T7} & \multicolumn{1}{|l|}{T8} & \multicolumn{1}{|l|}{average}\\ \hline
\boldmath\textbf{ExpertGate} & 1.54 & 1.41 & 2.51 & 1.69 & 0.9 & 1.58 & 1.13 & 0.91 & 1.46\\ \hline
\boldmath\textbf{Finetune} & 1.48 & 1.62 & 1.4 & 1.73 & 1.47 & 1.77 & 2.66 & 13.17 & 3.16\\ \hline
\boldmath\textbf{IMM} & 1.79 & 1.58 & 2.1 & 1.9 & 2.88 & 2.07 & 1.76 & 0.95 & 1.88\\ \hline
\boldmath\textbf{Int.Synapses} & 1.4 & 1.55 & 2.25 & 2.95 & 2.94 & 3.09 & 3 & 5.8 & 2.87\\ \hline
\boldmath\textbf{MAS} & 1.79 & 2.29 & 3.25 & 3.8 & 4.6 & 3.09 & 3.4 & 4.44 & 3.33\\ \hline
\boldmath\textbf{Joint} & 3.75 & 4.65 & 6.99 & 4.61 & 3.81 & 8.9 & 10.38 & 2.5 & 5.7\\ \hline
\end{tabular}
}
\vspace{0.18cm}
\caption{Few-shot ($\leq 10$) generalized Top 5 accuracy, Large-scale, Random Split} 
\label{tab:lsc_random_few_shot}
}
\vspace{-5mm}
\end{table}

\vspace{-4mm}
\section{Conclusions}
\label{sec:conclu}
\vspace{-2mm}
In this paper, we proposed two benchmarks to evaluate fact learning in a lifelong learning setup. A methodology was designed to split up an existing fact learning dataset into multiple tasks, taking the specific constraints into account and aiming for a setup that mimics real world application scenarios. With these benchmarks, we hope to foster research towards more large scale, human-like artificial visual learning systems and studying challenges like long-tail distribution.
\\ 

\noindent
{\bf Acknowledgements}
Rahaf Aljundi's research was funded by an FWO scholarship.
\section*{Appendix}
In the supplementary, we attach two folders that include the large-scale and mid-scale benchmarks annotations that we developed; see ``large-scale\_benchmarks'' and ``large-scale\_benchmarks''folders.  These folders have a comprehensive list of the tasks and the names of the facts of each of the Large-scale and mid-scale benchmarks.  This document also includes additional details and results, listed below.

\begin{enumerate}
\item {Mid-scale Task Similarities using average Word2vec space}
\item {Large Scale Semantic Splits 8 Tasks on word2vec space}
\item{Standard Accuracy (Long Tail/and Semantic/Random Improvement)}
\item{Long Tail Distribution Statistics on The Large Scale Dataset}
\item{SPO Generalization}
\item{Qualitative Examples}
\item{Mid-Scale dataset Dendogram}
\end{enumerate}

\section{Mid-scale Task Similarities using average Word2vec space (top-part) and geometric mean S,P, and O overlap  (bottom-part)}

\begin{table}
\centering
\vspace{-2mm}
\scalebox{0.7}
{
\begin{tabular}{@{}|l|r|r|r|r|l|r|r|r|r|@{}}  \hline
  \multicolumn{5}{|l|}{\boldmath\textbf{Semantic (0.02 word2vec mean similarity )}}  &  \multicolumn{5}{|l|}{\boldmath\textbf{Random (0.22 mean similarity )}}   \\ \hline
 & \multicolumn{1}{|l|}{\boldmath\textbf{T1}} & \multicolumn{1}{|l|}{\boldmath\textbf{T2}} & \multicolumn{1}{|l|}{\boldmath\textbf{T3}} & \multicolumn{1}{|l|}{\boldmath\textbf{T4}} &  & \multicolumn{1}{l}{\boldmath\textbf{T1}} & \multicolumn{1}{|l|}{\boldmath\textbf{T2}} & \multicolumn{1}{|l|}{\boldmath\textbf{T3}} & \multicolumn{1}{|l|}{\boldmath\textbf{T4}}\\  \hline
\boldmath\textbf{T1} & 1.00 & -0.16 & 0.04 & -0.55 & \boldmath\textbf{T1} & 1.00 & 0.16 & 0.39 & -0.31\\  \hline
\boldmath\textbf{T2} & -0.16 & 1.00 & -0.22 & -0.42 & \boldmath\textbf{T2} & 0.16 & 1.00 & -0.12 & -0.37\\  \hline
\boldmath\textbf{T3} & 0.04 & -0.22 & 1.00 & -0.5 & \boldmath\textbf{T3} & 0.39 & -0.12 & 1.00 & -0.01\\  \hline
\boldmath\textbf{T4} & -0.55 & -0.42 & -0.5 & 1.00 & \boldmath\textbf{T4} & -0.31 & -0.37 & -0.01 & 1.00\\    \hline  \hline
  \multicolumn{5}{|l|}{\boldmath\textbf{Semantic (0.25  g-mean of S,P, and O overlap )}}  &  \multicolumn{5}{|l|}{\boldmath\textbf{Random (0.84 g-mean of S,P, and O overlap)}}   \\ \hline
 & \multicolumn{1}{|l|}{\boldmath\textbf{T1}} & \multicolumn{1}{|l|}{\boldmath\textbf{T2}} & \multicolumn{1}{|l|}{\boldmath\textbf{T3}} & \multicolumn{1}{|l|}{\boldmath\textbf{T4}} &  & \multicolumn{1}{l}{\boldmath\textbf{T1}} & \multicolumn{1}{|l|}{\boldmath\textbf{T2}} & \multicolumn{1}{|l|}{\boldmath\textbf{T3}} & \multicolumn{1}{|l|}{\boldmath\textbf{T4}}\\  \hline
\boldmath\textbf{T1} & 1 & 0.0792 & 0.0774 & 0 & \boldmath\textbf{T1} & 1 & 0.19 & 0.23 & 0.12\\ \hline
\boldmath\textbf{T2} & 0.0792 & 1 & 0 & 0 & \boldmath\textbf{T2} & 0.19 & 1 & 0.19 & 0.19\\ \hline
\boldmath\textbf{T3} & 0.0774 & 0 & 1 & 0 & \boldmath\textbf{T3} & 0.23 & 0.19 & 1 & 0.13\\ \hline
\boldmath\textbf{T4}& 0 & 0 & 0 & 1 & \boldmath\textbf{T4} & 0.12 & 0.19 & 0.13 & 1\\ \hline

\end{tabular}

}
\caption{Mid scale Task Similarities using average Word2vec space (top-part) and geometric mean S,P, and O overlap  (bottom-part)}
\label{tab:mds_task_correlation}
\end{table}

\clearpage

\subsection{Large Scale Semantic Splits 8 Tasks on word2vec space}
 \begin{figure}[b!]
  \centering
    \includegraphics[width=0.8\textwidth]{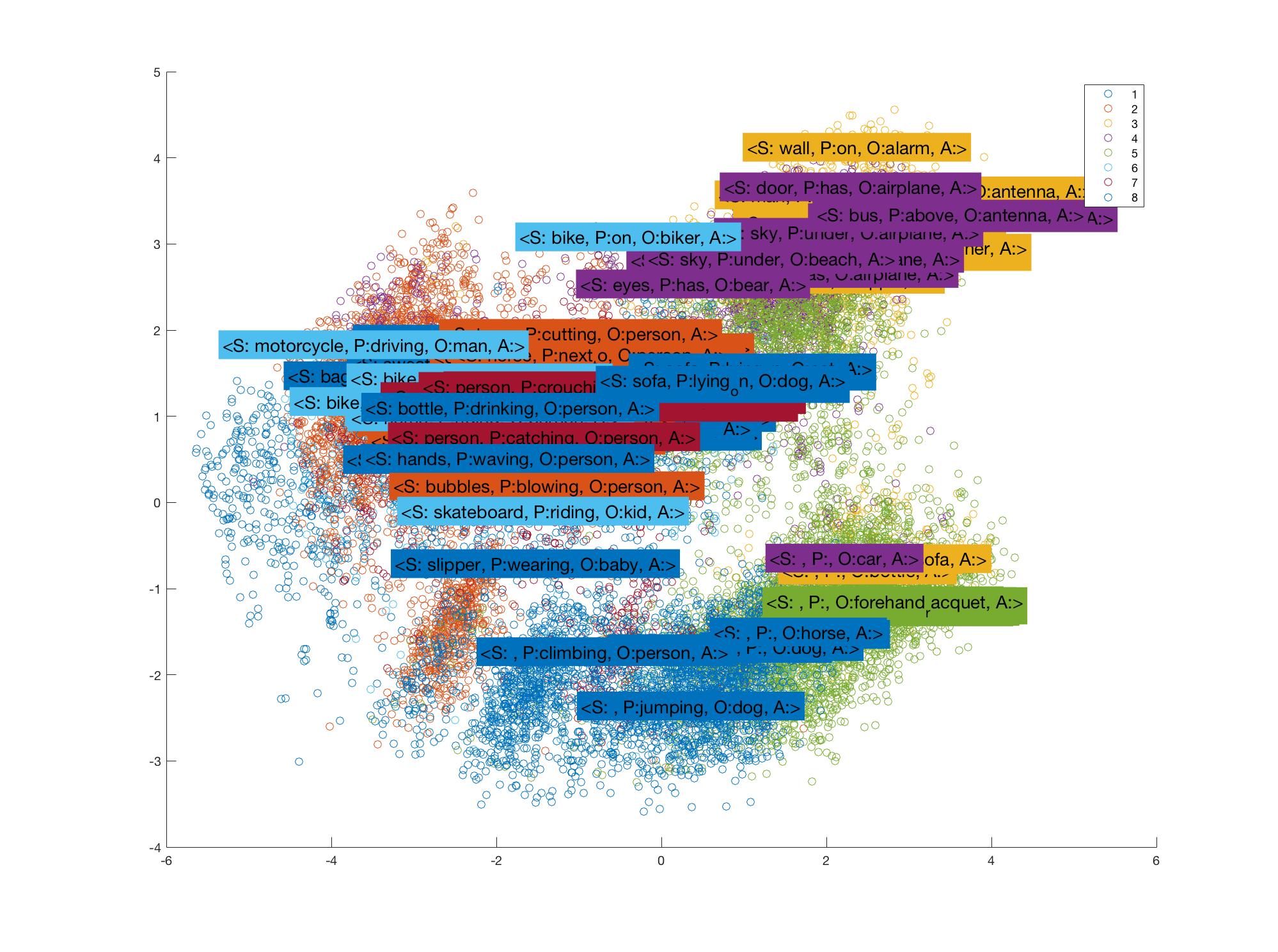}
      \caption{LSC Semantic Splits 8 Tasks on word2vec space (900 dimensions using PCA, each Task is color coded by a different color) }
      \label{fig_benchmark_vis}
\end{figure}

\clearpage

\subsection{Standard Accuracy (Long Tail/and Semantic/Random Improvement}

 \begin{figure*}[b!]
  \centering
   \includegraphics[width=0.38\textwidth]{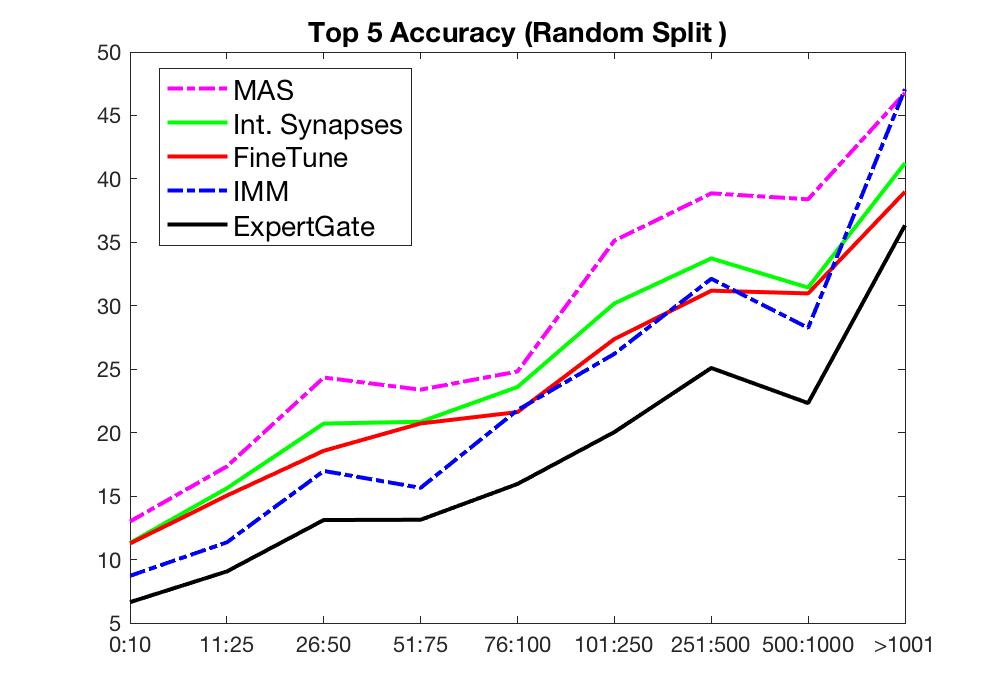}
    \includegraphics[width=0.38\textwidth]{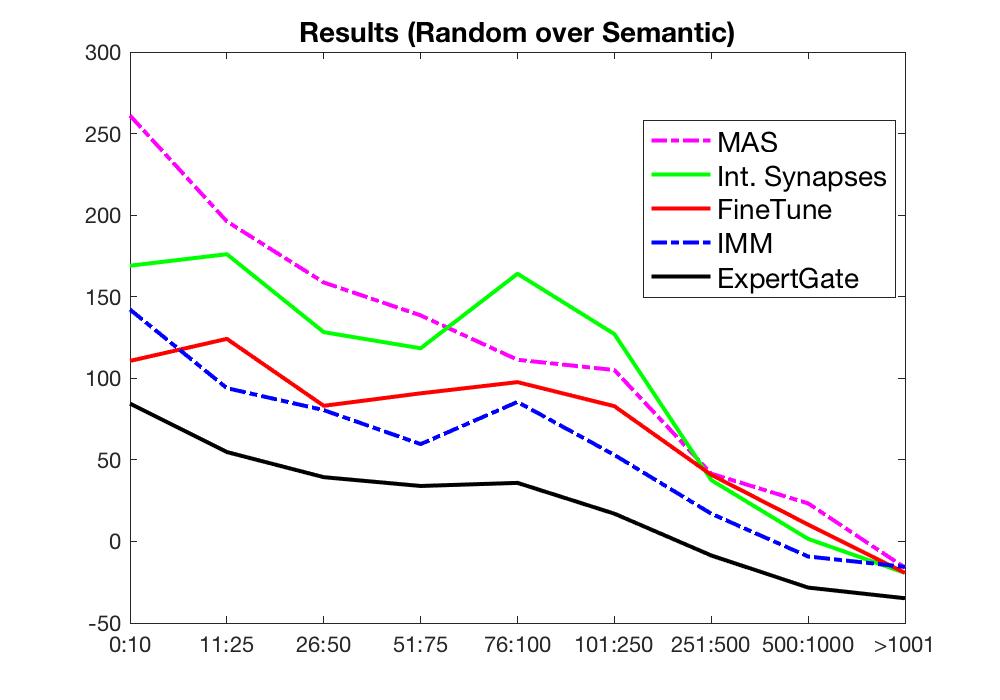}
     \includegraphics[width=0.38\textwidth]{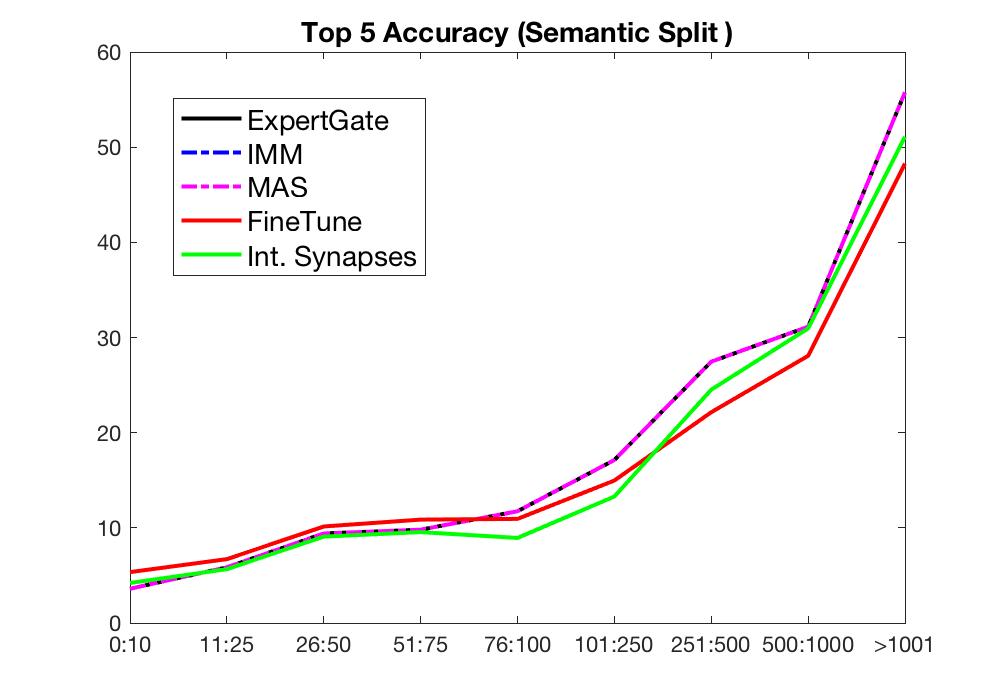}
          \includegraphics[width=0.38\textwidth]{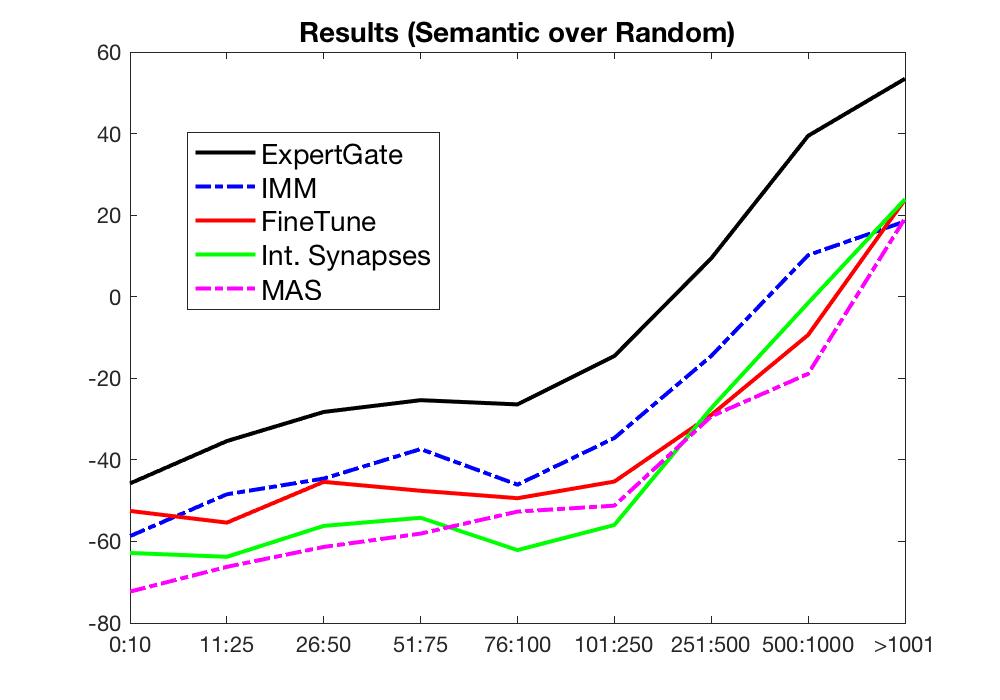}
            \vspace{-4mm}
      \caption{The x-axis in this figure shows the number of examples seen during training.  On the left: the y-axis shows the standard Top 5 Accuracy  for the random and the semantic splits. On the right: The y-axis shows the Random split improvement over the semantic split for each range and vice versa.}
      \label{fig_semantic_vs_random}
      \vspace{-4mm}
\end{figure*}

\clearpage

\pagebreak
\clearpage
\subsection{ SPO Generalization  }

It is desirable for each Life-long learning method to be able to generalize to understand an SPO interaction
from training examples involving its components, even when there are zero or very
few training examples for the exact SPO with all its parts S,P and O. For example, for $<$dog, riding, horse$>$ SPO  example $< \cdot$, riding, horse$>$ (the PO part) might have been seen more than 15 examples (TH=15) and $<$dog,$\cdot$ ,$\cdot>$  the S part might have been seen more than 15 examples.  Table ~\ref{tbl_gen1_15}  and  ~\ref{tbl_gen1_50} shows
the Top5 performance for SPOs for different LLL methods where the number of training examples is $\le$ 5 for generalization cases where SP$\ge$15,O$\ge TH$, or  P$\ge TH$,SO$\ge TH$, or    PO$\ge$15,S$\ge TH$, $TH=15$ and $TH-50$ for Tables ~\ref{tbl_gen1_15}  and  ~\ref{tbl_gen1_50}, respectively. Similarly, Table~\ref{tbl_gen2_15} and ~\ref{tbl_gen2_50}  shows a different set of generalization cases which are  SP$\ge TH$,PO$\ge$15,SO$\ge TH$ or SP$\ge TH$,PO$> TH$  or SP$\ge TH$,SO$\ge TH$ or   PO$\ge$15,SO$\ge TH$, $TH=15$ and $TH-50$ for Table~\ref{tbl_gen2_15} and ~\ref{tbl_gen2_50} respectively.

\begin{table}[h!]
\centering
\begin{tabular}{|l|l|l|l|l|}
\hline \textbf{SPO Generalization} &   SP$\ge$15,O$\ge$15  & P$\ge$15,SO$\ge$15&  PO$\ge$15,S$\ge$15 \\ \hline
Finetuning        &  0.1011 &  0.2744  & 0.1235 \\ \hline
Int. Synapses     & 0.1041 &    0.2111 &   0.0617\\ \hline
MAS               & 0.0681  &   0.0950 & 0.0309 \\ \hline
\end{tabular}
\caption{SPO (Interaction) Generalization (Top5 Performance): The entire entire action has been seen very rare  (SPO$\le$5 examples in this table) but individual pieces has been seen (like SP and O, PO and S, or SO and P), TH=15 in this table. For example, for $<$dog, riding, horse$>$ SPO  example $< \cdot$, riding, horse$>$ (the PO part) might have been seen more than 15 examples (TH=15) and $<$dog,$\cdot$ ,$\cdot>$  the S part might have been seen more than 15 examples (TH=15) }
\label{tbl_gen1_15}
\end{table}

\begin{table}[h!]
\centering
\begin{tabular}{|l|l|l|l|l|}
\hline \textbf{SPO Generalization} &   SP$\ge$50,O$\ge$50  & P$\ge$50,SO$\ge$50&  PO$\ge$50,S$\ge$50 \\
Finetuning        &  0.1725 &  0.3156  & 0.0866 \\ \hline
Int. Synapses     & 0.1258 &    0.2018 &   0.0765\\ \hline
MAS               &  0.1002  &   0.2328 & 0.0561\\ \hline
\end{tabular}
\caption{SPO (Interaction) Generalization (Top5 Performance): The entire entire action has been seen very rare  (SPO$\le$5 examples in this table) but individual pieces has been seen (like SP and O, PO and S, or SO and P), TH=50 in this table.}
\label{tbl_gen1_50}
\end{table}

\begin{table}[h!]
\centering
\begin{tabular}{|l|l|l|l|l|}
\hline\textbf{SPO Generalization} & SP$\ge$15,PO$\ge$15,SO$>$15 &    SP$\ge$15,PO$>$15  & SP$\ge$15,SO$\ge$15&  PO$\ge$15,SO$\ge$15 \\ \hline
Finetuning        & 0.2091 &  0.2937  &0.2091& 0.3380 \\ \hline
Int. Synapses     & 0.1052 &   0.1281& 0.1121& 0.2161\\ \hline
MAS               &  0.1442  &  0.1957 & 0.1415& 0.2161 \\ \hline
\end{tabular}
\caption{SPO (Interaction) Generalization (Top5 Performance): The entire entire action has been seen very rare  (SPO$\le$5 examples in this table) but individual pieces has been seen (like SP, PO, or SO), TH=15 TH=15.}
\label{tbl_gen2_15}
\end{table}

\begin{table}[h!]
\centering
\label{my-label}
\begin{tabular}{|l|l|l|l|l|}
\hline \textbf{SPO Generalization} & SP$\ge$50,PO$\ge$50,SO$\ge$50  &    SP$\ge$50,PO$ge$50  &   SP$\ge$50,SO$\ge$50    & PO$\ge$50,SO$\ge$50\\ \hline
Finetuning        & 0.2331 & 0.2866 & 0.2298& 0.3329\\ \hline
Int. Synapses     & 0.107 & 0.1390 &  0.1128& 0.1949\\ \hline
MAS               &  0.1478 & 0.1856  & 0.1446& 0.2190\\ \hline
\end{tabular}
\caption{SPO (Interaction) Generalization: The entire entire action has been seen very rare  (SPO$\le$5 examples in this table) but individual pieces has been seen (like SP, PO, or SO) , TH=50.}
\label{tbl_gen2_50}
\end{table}

\clearpage
\subsection{Qualitative Examples }
This section shows correctly and incorrectly classified examples for each of fine-tuning, Intl. Synapsses, and Memory Aware Synapses.
\begin{figure*}[h!]
  \centering
    \includegraphics[width=0.40\textwidth]{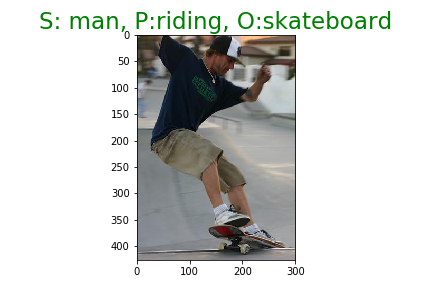} \includegraphics[width=0.40\textwidth]{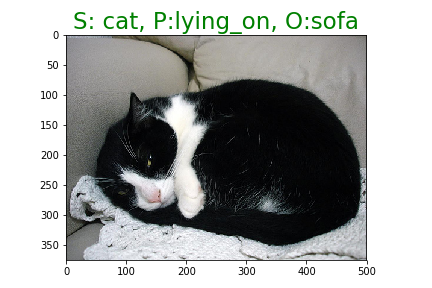}
 \includegraphics[width=0.40\textwidth]{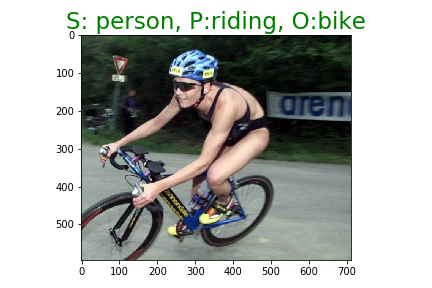}
  \includegraphics[width=0.40\textwidth]{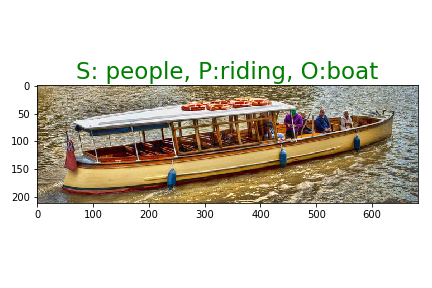}
   \includegraphics[width=0.40\textwidth]{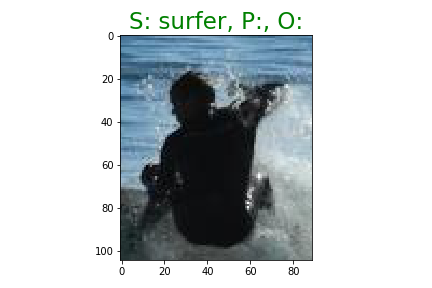}
     \includegraphics[width=0.40\textwidth]{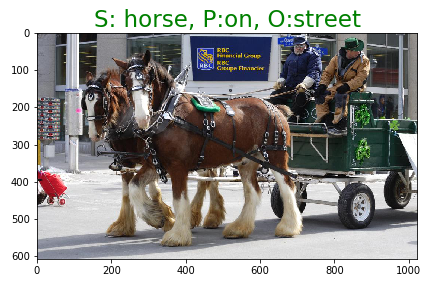}
	           \includegraphics[width=0.40\textwidth]{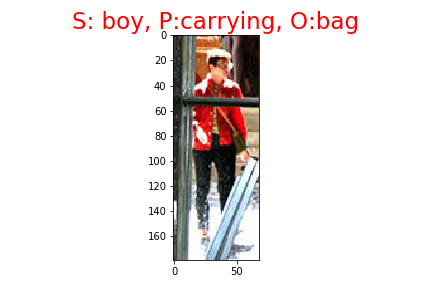}
                \includegraphics[width=0.40\textwidth]{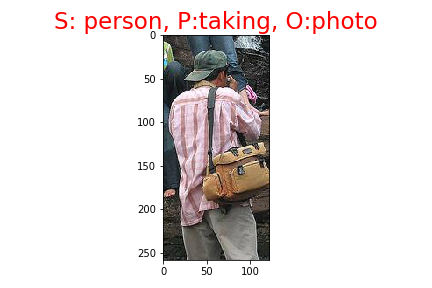}
      \caption{Correctly (title in green) and incorrectly (title in red) classified examples by MAS~\cite{mas} on our Large-scale LLFL benchmark.}
      \label{fig_Qual_positive_MAS}
\end{figure*}

\begin{figure*}[h!]
  \centering
    \includegraphics[width=0.45\textwidth]{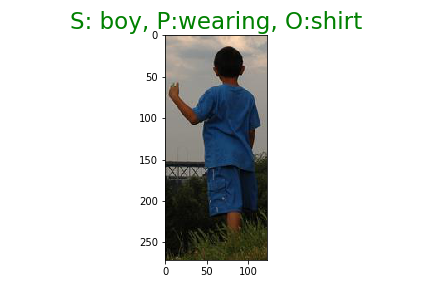} \includegraphics[width=0.45\textwidth]{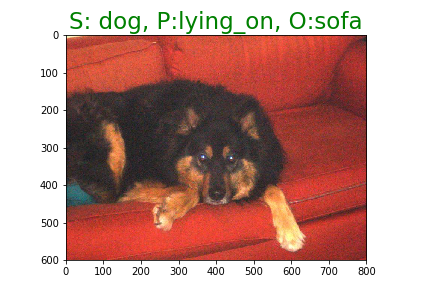}
 \includegraphics[width=0.45\textwidth]{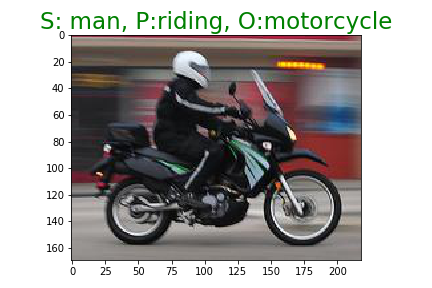}
  \includegraphics[width=0.45\textwidth]{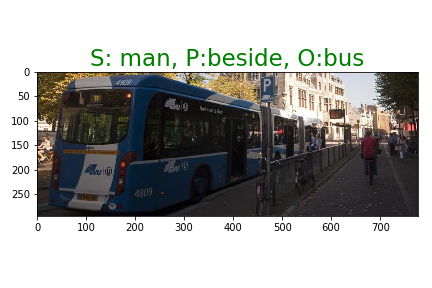}
   \includegraphics[width=0.45\textwidth]{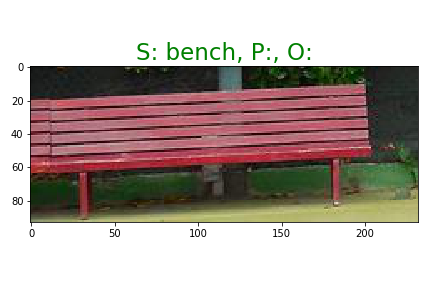}
	           \includegraphics[width=0.40\textwidth]{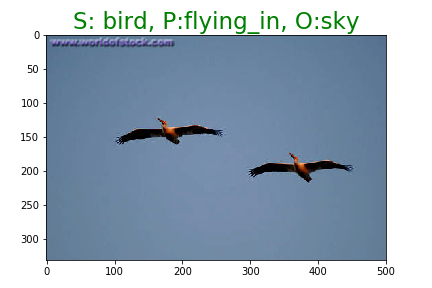}
               \includegraphics[width=0.45\textwidth]{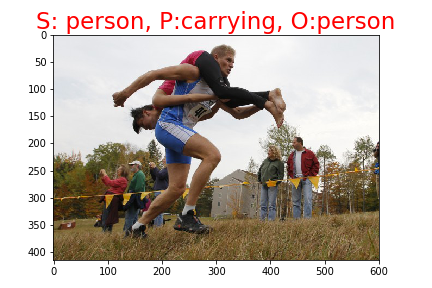}
	           \includegraphics[width=0.40\textwidth]{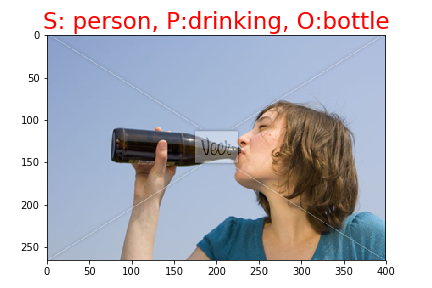}
       \caption{Correctly (title in green) and incorrectly (title in red) classified examples by Intelligent Synapses~\cite{zenke2017improved} on our Large-scale LLFL benchmark.}
      \label{fig_Qual_positive_SI}
\end{figure*}

\begin{figure*}[h!]
  \centering
    \includegraphics[width=0.45\textwidth]{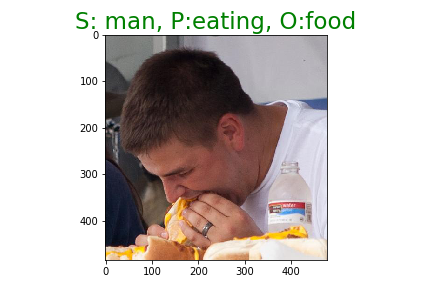} \includegraphics[width=0.45\textwidth]{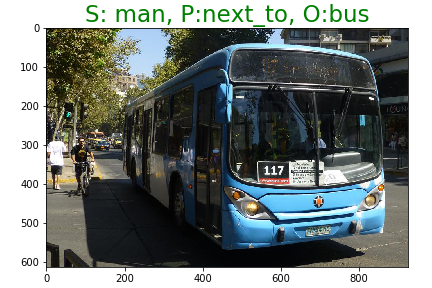}
 \includegraphics[width=0.45\textwidth]{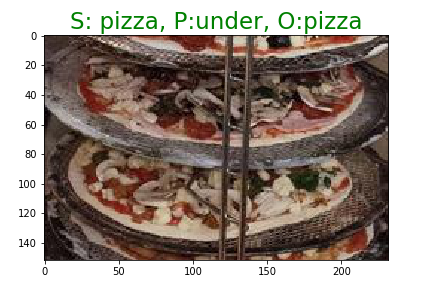}
  \includegraphics[width=0.45\textwidth]{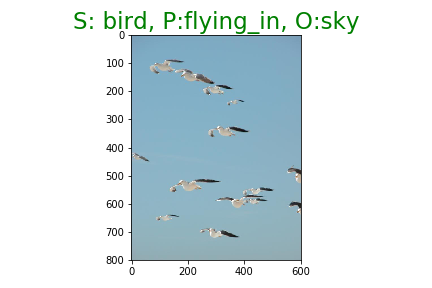}
   \includegraphics[width=0.45\textwidth]{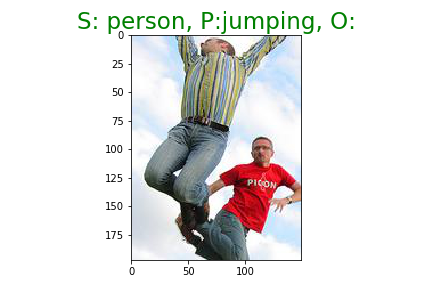}
     \includegraphics[width=0.45\textwidth]{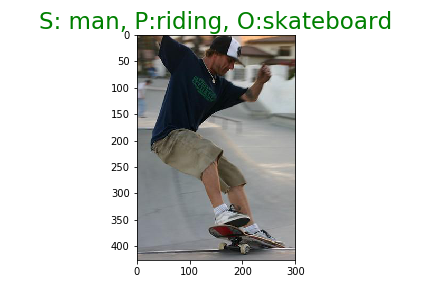}
	           \includegraphics[width=0.40\textwidth]{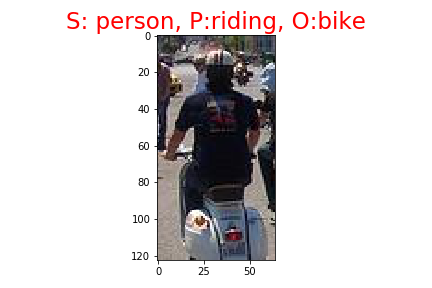}
               \includegraphics[width=0.40\textwidth]{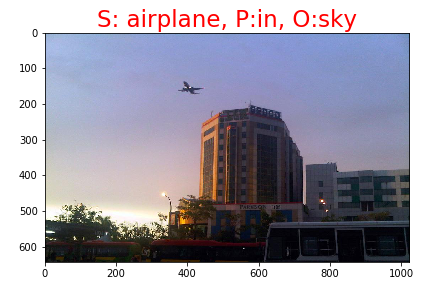}
      \caption{Correctly (title in green) and incorrectly (title in red) Classified examples by Finetuning on our Large-scale LLFL benchmark.}
      \label{fig_Qual_positive_F}
\end{figure*}

\pagebreak
\clearpage
\section{ Mid-Scale benchmark Dendogram}

\begin{figure*}[h!]
  \centering
    \includegraphics[width=1.0\textwidth]{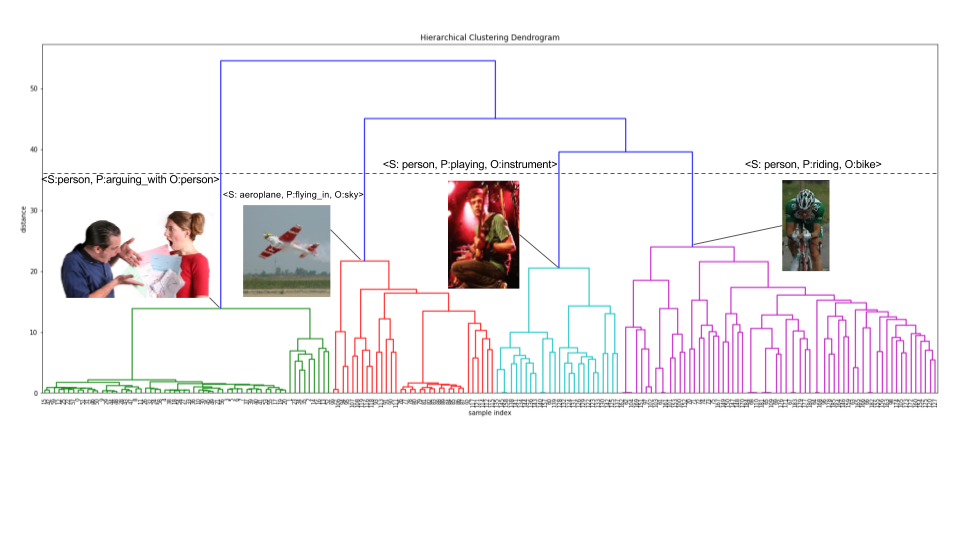}
      \caption{Lifelong Fact Learning for the Midscale dataset}
      \label{fig_6DS}
\end{figure*}

Figure \ref{fig_6DS} shows the dendogram obtained from the agglomerative clustering performed in the word2vec space of the facts from the mid-scale dataset. The different colors indicate the different clusters. Each cluster later forms a task. 
\begin{enumerate}
\item Cluster in {\color{magenta}magenta} that mostly represents person actions and contains the person fact that is needed in the rest of that tasks.
\item Cluster in {\color{red}red} resembles the second task and is mainly composed of facts of different objects.
\item Cluster in {\color{cyan}Cyan} is the third cluster  which contains facts describing humans holding or  playing musical instruments.
\item Cluster in {\color{green}Green} (last) is composed of the fact belonging to the green cluster  that is composed of facts describing human interactions.
\end{enumerate}

\bibliographystyle{splncs03}
\bibliography{egbib}

\begin{thebibliography}{10}
\providecommand{\url}[1]{\texttt{#1}}
\providecommand{\urlprefix}{URL }

\bibitem{mas}
Aljundi, R., Babiloni, F., Elhoseiny, M., Rohrbach, M., Tuytelaars, T.: Memory
  aware synapses: Learning what (not) to forget. In: ECCV (2018)

\bibitem{aljundi2016expert}
Aljundi, R., Chakravarty, P., Tuytelaars, T.: Expert gate: Lifelong learning
  with a network of experts. In: CVPR (2017)

\bibitem{chao2016empirical}
Chao, W.L., Changpinyo, S., Gong, B., Sha, F.: An empirical study and analysis
  of generalized zero-shot learning for object recognition in the wild. In:
  European Conference on Computer Vision. pp. 52--68. Springer (2016)

\bibitem{chaudhry2018riemannian}
Chaudhry, A., Dokania, P.K., Ajanthan, T., Torr, P.H.: Riemannian walk for
  incremental learning: Understanding forgetting and intransigence. In:
  International Conference on Machine Learning (2018)

\bibitem{chen2013neil}
Chen, X., Shrivastava, A., Gupta, A.: Neil: Extracting visual knowledge from
  web data. In: Computer Vision (ICCV), 2013 IEEE International Conference on.
  pp. 1409--1416. IEEE (2013)

\bibitem{divvala2014learning}
Divvala, S.K., Farhadi, A., Guestrin, C.: Learning everything about anything:
  Webly-supervised visual concept learning. In: Proceedings of the IEEE
  Conference on Computer Vision and Pattern Recognition. pp. 3270--3277 (2014)

\bibitem{elhoseiny2017sherlock}
Elhoseiny, M., Cohen, S., Chang, W., Price, B.L., Elgammal, A.M.: Sherlock:
  Scalable fact learning in images. In: AAAI. pp. 4016--4024 (2017)

\bibitem{fernando2017pathnet}
Fernando, C., Banarse, D., Blundell, C., Zwols, Y., Ha, D., Rusu, A.A.,
  Pritzel, A., Wierstra, D.: Pathnet: Evolution channels gradient descent in
  super neural networks. arXiv preprint arXiv:1701.08734  (2017)

\bibitem{gong2014multi}
Gong, Y., Ke, Q., Isard, M., Lazebnik, S.: A multi-view embedding space for
  modeling internet images, tags, and their semantics. International journal of
  computer vision  106(2),  210--233 (2014)

\bibitem{he2016deep}
He, K., Zhang, X., Ren, S., Sun, J.: Deep residual learning for image
  recognition. In: Proceedings of the IEEE conference on computer vision and
  pattern recognition. pp. 770--778 (2016)

\bibitem{kading2016fine}
K{\"a}ding, C., Rodner, E., Freytag, A., Denzler, J.: Fine-tuning deep neural
  networks in continuous learning scenarios. In: Asian Conference on Computer
  Vision. pp. 588--605. Springer (2016)

\bibitem{kirkpatrick2016overcoming}
Kirkpatrick, J., Pascanu, R., Rabinowitz, N., Veness, J., Desjardins, G., Rusu,
  A.A., Milan, K., Quan, J., Ramalho, T., Grabska-Barwinska, A., et~al.:
  Overcoming catastrophic forgetting in neural networks. Proceedings of the
  national academy of sciences p. 201611835 (2017)

\bibitem{kiros2014unifying}
Kiros, R., Salakhutdinov, R., Zemel, R.S.: Unifying visual-semantic embeddings
  with multimodal neural language models. arXiv preprint arXiv:1411.2539
  (2014)

\bibitem{krizhevsky2012imagenet}
Krizhevsky, A., Sutskever, I., Hinton, G.E.: Imagenet classification with deep
  convolutional neural networks. In: Advances in neural information processing
  systems. pp. 1097--1105 (2012)

\bibitem{lee2017overcoming}
Lee, S.W., Kim, J.H., Jun, J., Ha, J.W., Zhang, B.T.: Overcoming catastrophic
  forgetting by incremental moment matching. In: Advances in Neural Information
  Processing Systems. pp. 4652--4662 (2017)

\bibitem{li2016learning}
Li, Z., Hoiem, D.: Learning without forgetting. In: European Conference on
  Computer Vision. pp. 614--629. Springer (2016)

\bibitem{lomonaco17corl}
Lomonaco, V., Maltoni, D.: Core50: a new dataset and benchmark for continuous
  object recognition. In: Conference on Robot Learning (2017)

\bibitem{lopez2017gradient}
Lopez-Paz, D., Ranzato, M.: Gradient episodic memory for continual learning.
  In: Advances in Neural Information Processing Systems (2017)

\bibitem{mikolov2013distributed}
Mikolov, T., Sutskever, I., Chen, K., Corrado, G.S., Dean, J.: Distributed
  representations of words and phrases and their compositionality. In: Advances
  in neural information processing systems. pp. 3111--3119 (2013)

\bibitem{mitchell2015never}
Mitchell, T.M., Cohen, W.W., Hruschka~Jr, E.R., Talukdar, P.P., Betteridge, J.,
  Carlson, A., Mishra, B.D., Gardner, M., Kisiel, B., Krishnamurthy, J.,
  et~al.: Never ending learning. In: AAAI. pp. 2302--2310 (2015)

\bibitem{Pasquale2016IROS}
Pasquale, G., Ciliberto, C., Rosasco, L., Natale, L.: Object identification
  from few examples by improving the invariance of a deep convolutional neural
  network. In: 2016 IEEE/RSJ International Conference on Intelligent Robots and
  Systems (IROS). pp. 4904--4911 (Oct 2016),
  \url{http://ieeexplore.ieee.org/document/7759720/}

\bibitem{plummer2017phrase}
Plummer, B.A., Mallya, A., Cervantes, C.M., Hockenmaier, J., Lazebnik, S.:
  Phrase localization and visual relationship detection with comprehensive
  image-language cues. In: Proceedings of the IEEE Conference on Computer
  Vision and Pattern Recognition. pp. 1928--1937 (2017)

\bibitem{rebuffi2016icarl}
Rebuffi, S.A., Kolesnikov, A., Lampert, C.H.: icarl: Incremental classifier and
  representation learning. arXiv preprint arXiv:1611.07725  (2016)

\bibitem{romera2015embarrassingly}
Romera-Paredes, B., Torr, P.: An embarrassingly simple approach to zero-shot
  learning. In: International Conference on Machine Learning. pp. 2152--2161
  (2015)

\bibitem{shmelkov17iccv}
Shmelkov, K., Schmid, C., Alahari, K.: Incremental learning of object detectors
  without catastrophic forgetting. In: The IEEE International Conference on
  Computer Vision (ICCV) (2017)

\bibitem{simonyan2014very}
Simonyan, K., Zisserman, A.: Very deep convolutional networks for large-scale
  image recognition. arXiv preprint arXiv:1409.1556  (2014)

\bibitem{thrun1998clustering}
Thrun, S., O’Sullivan, J.: Clustering learning tasks and the selective
  cross-task transfer of knowledge. In: Learning to learn, pp. 235--257.
  Springer (1998)

\bibitem{triki2017encoder}
Triki, A.R., Aljundi, R., Blaschko, M.B., Tuytelaars, T.: Encoder based
  lifelong learning. arXiv preprint arXiv:1704.01920  (2017)

\bibitem{wang2017growing}
Wang, Y.X., Ramanan, D., Hebert, M.: Growing a brain: Fine-tuning by increasing
  model capacity. In: IEEE Computer Society Conference on Computer Vision and
  Pattern Recognition (CVPR) (2017)

\bibitem{zenke2017improved}
Zenke, F., Poole, B., Ganguli, S.: Continual learning through synaptic
  intelligence. In: Proceedings of the 34th International Conference on Machine
  Learning. vol.~70, pp. 3987--3995. PMLR (06--11 Aug 2017)

\bibitem{zhang2018large}
Zhang, J., Kalantidis, Y., Rohrbach, M., Paluri, M., Elgammal, A., Elhoseiny,
  M.: Large-scale visual relationship understanding. arXiv preprint
  arXiv:1804.10660  (2018)

\end{thebibliography}

\end{document}